\documentclass[sigconf,nonacm,balance=false]{acmart}

\usepackage{adjustbox}
\usepackage{microtype}
\usepackage{float}
\usepackage{dblfloatfix}
\usepackage{fancyvrb}
\usepackage{fvextra}
\usepackage{placeins}
\usepackage{multirow}
\usepackage{tabularx}
\usepackage{longtable}
\usepackage{makecell}
\usepackage{enumitem}
\usepackage{xspace}
\usepackage{algorithm}
\usepackage{algpseudocode}
\usepackage{mathtools}
\usepackage{pdfpages}
\usepackage[normalem]{ulem}
\usepackage[capitalize,noabbrev]{cleveref}

\usepackage{etoc}

\newcounter{TraceBenchSavedDblTopNumber}
\newcolumntype{Y}{>{\centering\arraybackslash}X}
\newcolumntype{S}{>{\raggedright\arraybackslash}p{1.8cm}}
\newcolumntype{M}{>{\raggedright\arraybackslash}p{3.2cm}}

\makeatletter
\DeclareRobustCommand\onedot{\futurelet\@let@token\@onedot}
\def\@onedot{\ifx\@let@token.\else.\null\fi\xspace}

\makeatother

\makeatletter
\newcommand{\TraceBenchQuietVerbatimSpacing}{%
  \let\@vspace\@vspace@orig
  \let\@vspacer\@vspacer@orig
}
\makeatother

\theoremstyle{plain}

\theoremstyle{definition}

\theoremstyle{remark}

\newcommand{\name}{TraceBench\xspace}
\newcommand{\papertitle}{\name{}: Controlled Evaluation of LLM Agents for Time-Series Root-Cause Attribution}
\providecommand{\code}[1]{%
  \begingroup
  \def\UrlFont{\ttfamily}%
  \nolinkurl{#1}%
  \endgroup
}
\providecommand{\titlecode}[1]{\code{#1}}
\DeclareRobustCommand{\termSimulator}{\code{simulator}\xspace}

\title{\papertitle}
\author{Tommaso Bendinelli}
\affiliation{%
  \institution{Department of Computer Science}
  \institution{ETH Zürich, Switzerland}
  \city{}
  \country{}}
\affiliation{%
  \institution{CSEM SA, Switzerland}
  \city{}
  \country{}
}

\author{Artur Dox}
\affiliation{%
  \institution{Independent Researcher}
  \city{}
  \country{}
}
\email{}

\author{Christian Holz}
\affiliation{%
  \institution{Department of Computer Science}
  \institution{ETH Zürich, Switzerland}
  \city{}
  \country{}
}

\begin{document}
\etocdepthtag.toc{tracebench-main}

\begin{abstract}
LLM agents are increasingly applied to anomaly detection and root-cause analysis in time-series observations collected from real-world systems; however, their performance on these tasks has not been systematically evaluated under controlled conditions.
We introduce \name{}, a simulation-based framework for generating controlled root-cause attribution tasks.
In each generated task, an agent receives time-series observations produced by simulating a physical dynamical system and must determine whether a system parameter was altered during the simulation and, if so, which one.
Using \name{}, we generate tasks from three interpretable mechanical systems and systematically evaluate four LLM agents across controlled experimental conditions, yielding new insights into how these agents analyze time-series observations from dynamical systems.
Our results show that agents benefit substantially from domain context and explore data primarily through numerical console output rather than visualizations.
We also find that agents generally perform worse when required to produce a Python script that maps each time-series sample to a predicted root-cause label than when they submit predictions directly.
We release our datasets, agent trajectories, experimental results, and a leaderboard on our website, \href{https://tracebench.github.io}{tracebench.github.io}.
\end{abstract}


\maketitle
\pdfinfo{/Author (Tommaso Bendinelli, Artur Dox, Christian Holz)}
\phantomsection\label{docstart:kdd_main}

\section{Introduction}
Real-world root-cause analysis requires combining observed data with knowledge of the system that generated them~\citep{isermann2005modelbased}. 
For example, an engineer investigating the cause of abnormal vibration traces from a machine must reason jointly about the observed measurements and the system that produced them.
This interplay between data exploration and domain knowledge provides a useful setting for evaluating whether LLM agents can integrate system descriptions with iterative, tool-assisted time-series analysis.

A large body of work evaluates LLM agents on machine-learning engineering and data-analysis tasks~\citep{chan2024mlebench,hu2024infidabench,sahu2024insightbench,jing2024dsbench,cai2025timeseriesgym}, with a subset focusing on root-cause analysis over software telemetry, fault-injected cloud environments, and industrial asset-management workflows~\citep{xu2025openrca,chen2025aiopslab,patel2025assetopsbench}.
These benchmarks evaluate agents in realistic settings that require combining domain knowledge with data inspection, and therefore provide valuable evidence about end-to-end diagnostic capability. 
However, they do not systematically isolate an agent's ability to combine temporal and physical reasoning over multivariate time series with process-specific knowledge. 

Existing language–time-series datasets span a broad range of reasoning tasks, but are not designed for controlled agentic root-cause attribution. 
They typically pair observations with questions, sample-specific or timestamp-aligned text, visualizations, or descriptions of visible patterns~\citep{merrill2024language,yang2026timera,liu2024timemmd,kim2024multimodalforecaster}, rather than with detailed mechanistic descriptions of the systems that generated the observations. 
Consequently, they do not directly test whether an LLM agent can use system-specific knowledge to formulate and refine diagnostic hypotheses through multi-step, tool-assisted analysis. 
In this paper, we introduce \name{}, a simulation-based framework designed to generate controlled root-cause attribution tasks and systematically study these aspects.

\begin{figure*}[t]
\centering
\includegraphics[width=\textwidth]{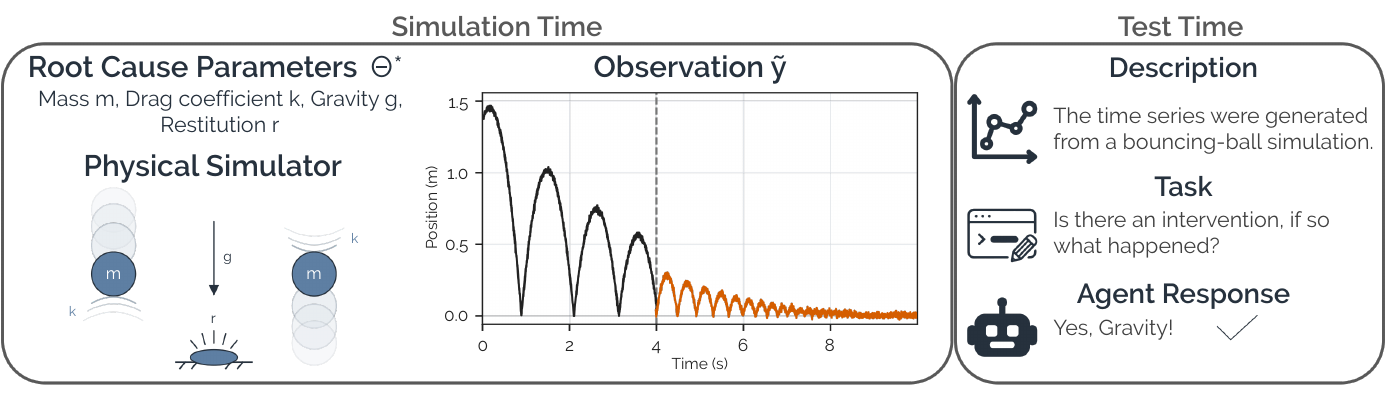}
\Description{\name{} pipeline: a physical simulator produces multivariate time series before and after a parameter intervention; the low-noise BallDrop position observation is black before and orange after a dashed gravity-intervention marker; the agent identifies the changed root-cause parameter.}
\caption{Overview of the \name{} framework using a bouncing-ball simulator.
In this example, the black and orange segments show the position observation under the low-noise setting before and after an intervention at $t_{\mathrm{int}}=4.0$.
The intervention increases gravitational acceleration, and the dashed line marks the intervention time.
At test time, the agent determines from the observed time-series channels whether an intervention occurred and, if so, which parameter changed.
}
\label{fig:data_generation}
\end{figure*}

Consider the following illustrative task:
Suppose we drop a ball from a specified height with a specified initial velocity and observe its height and velocity.
The ball's motion follows a trajectory governed by parameters such as the coefficient of restitution, the ball's mass, the drag coefficient, and gravitational acceleration.
Now suppose we intervene by modifying the value of one of these parameters while the ball is still bouncing.
This intervention changes the ball's motion and is reflected in the resulting time-series observations.
We ask the agent to determine whether a parameter was modified and, if so, which parameter accounts for the observed change in trajectory.
\Cref{fig:data_generation} illustrates this example and provides an overview of the \name{} framework.

Using \name{}, we systematically evaluate four LLM agents under controlled conditions that vary observation noise, access to descriptions of the physical system and measured signals, and the availability of labeled time-series examples with known root causes.
We further compare direct-answer submission, in which agents provide predictions directly, with programmatic submission, in which they produce a Python script that maps each time-series sample to a predicted root-cause label.
We deliberately evaluate agents across three interpretable mechanical systems whose behavior can be understood using basic physics, thereby limiting dependence on specialized domain expertise.
The resulting tasks test whether agents can interpret time-series observations in light of textual system descriptions, identify deviations from expected dynamics, and infer the mechanisms most consistent with the observed evidence.

\begin{figure}[t]
\includegraphics[width=\linewidth]{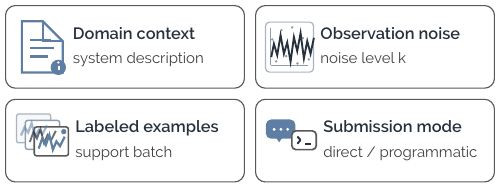}
\Description{Four cards summarize the independently controlled \name{} task axes: domain context, observation noise, labeled examples, and direct or programmatic submission.}
\caption{The four controllable axes in \name{}.}
\label{fig:controllable_task_axes}
\end{figure}

In summary, our contributions are the following:
First, we introduce \name{}, a simulation-based framework for generating root-cause attribution tasks along four controllable axes: domain context, observation noise, labeled examples, and submission mode, as summarized in \Cref{fig:controllable_task_axes}.
This controlled setting isolates whether an agent can combine temporal and physical reasoning over multivariate time-series observations with process-specific knowledge.
Second, we instantiate the framework with three mechanical systems and systematically evaluate four LLM agents in both direct-answer and programmatic submission settings.
Third, we analyze performance and agent behavior and find that agents rely primarily on console output rather than visualizations and perform worse when required to produce a Python script than when submitting predictions directly.
We release the complete benchmark, generated datasets, agent trajectories, and a public leaderboard at \href{https://tracebench.github.io}{tracebench.github.io}.

\FloatBarrier

\section{Related Work}

\label{sec:related_work}
Our work studies agentic root-cause analysis of multivariate time-series observations accompanied by system-specific textual descriptions. 
We position \name{} relative to two main lines of work: (i) agentic systems and benchmarks for time-series analysis and diagnosis, and (ii) LLM benchmarks, datasets, and models for time-series understanding and reasoning with textual information, typically evaluated in single-turn settings without iterative tool use.

\subsection{Agentic benchmarks for time-series analysis and diagnosis}
Recent work evaluates systems on time-series tasks that require analytical decomposition, the use of computational tools, or iterative reasoning based on contextual information and execution feedback.
TimeSeriesGym~\citep{cai2025timeseriesgym} evaluates agents on machine-learning engineering tasks that involve time series and require them to produce multiple types of executable artifacts.
TS-Reasoner~\citep{ye2026tsreasoner} studies program-aided decomposition of compositional time-series inference tasks, while TSAIA~\citep{ye2025tsaia} evaluates the construction of multi-step analytical workflows drawn from scientific and engineering applications.
TemporalBench~\citep{weng2026temporalbench} evaluates historical understanding, contextual reasoning, and event-conditioned prediction across multiple domains, including physical systems.
A complementary line of work studies agentic root-cause analysis in operational and industrial environments.
OpenRCA~\citep{xu2025openrca} benchmarks root-cause localization from heterogeneous software telemetry, while AIOpsLab~\citep{chen2025aiopslab} evaluates agents in interactive, fault-injected cloud environments.
AssetOpsBench~\citep{patel2025assetopsbench} broadens agent evaluation to industrial monitoring and maintenance workflows.

These works share \name{}'s emphasis on multi-step, context-informed analysis and diagnosis. 
They differ, however, in evaluation scope: prior benchmarks address a broader range of analytical and operational tasks, often involving heterogeneous data, artifacts, or application-specific workflows, whereas \name{} isolates root-cause attribution from multivariate time-series observations using controlled physical simulations with intervention-defined ground truth. 
Each sample corresponds either to no intervention or to a single change in one candidate simulator parameter, yielding a known ground-truth label by construction.
This controlled setup enables systematic analysis of agent behavior by independently varying four task axes: domain context, observation noise, labeled examples, and submission mode.

\subsection{LLM benchmarks and datasets for time-series reasoning with textual information}
A large body of work probes time-series understanding and reasoning in single-turn settings without tool use.
\citet{merrill2024language} evaluate etiological reasoning, factual question answering, and context-aided forecasting; they also test whether a model can identify the scenario that most plausibly generated an observed series.
\citet{fons2024evaluating} introduce a taxonomy and benchmark for time-series feature understanding, while TimeSeriesExam~\citep{cai2024timeseriesexam} evaluates pattern recognition, noise understanding, similarity, anomaly detection, and causal reasoning.
TIME-RA~\citep{yang2026timera} more directly studies anomaly diagnosis by combining raw time series, domain context, visualizations, fine-grained anomaly categories, and generated explanations.
\citet{gruver2024large} demonstrate that LLMs can act as zero-shot time-series forecasters, while \citet{tan2024are} analyze when language-model priors help or hurt forecasting.
SciTS~\citep{anonymous2025scits} studies scientific time-series understanding and generation with LLMs.
Time-MQA~\citep{kong2025timemqa} introduces multitask question answering for time series and the TSQA dataset.
TRQA~\citep{jing2025trqa} evaluates question answering for time-series reasoning across multiple task families, including anomaly detection, classification, characterization, comparison, transformation, and reasoning about temporal relationships.
ITFormer~\citep{wang2025itformer} proposes datasets and model interfaces for temporal-textual question answering, while ChatTS~\citep{xie2024chatts} demonstrates that synthetic alignment data can improve multimodal time-series understanding and reasoning.
Time-MMD~\citep{liu2024timemmd} aligns numerical time series with textual information for multimodal time-series analysis.
The TimeText Corpus, introduced alongside Multi-Modal Forecaster~\citep{kim2024multimodalforecaster}, provides timestamp-aligned textual and numerical data for forecasting.
BEDTime~\citep{sen2025bedtime} evaluates the automatic description of time series through recognition, differentiation, and natural-language generation tasks.

These resources provide important benchmarks linking signals and language, but their text is typically tied to the observed sample—for example, as a question, a timestamp-aligned annotation, or a description of visible patterns. 
\name{} instead provides a sample-independent, mechanistic description of the data-generating system: what is measured, which parameters may change, and the physical dynamics governing the system. 
The benchmark therefore evaluates whether agents can use this prior system description to formulate hypotheses about the effects of candidate parameter changes and test those hypotheses against the observed trajectories.

\FloatBarrier
\section{\name{} Framework}
\label{sec:overview}
\name{} is a simulation-based framework for generating controlled agentic root-cause attribution tasks.
Each \name{} task presents an agent with a test batch of time-series samples generated by simulating a dynamical system. 
The agent must, for each sample, identify which simulator parameter, if any, has changed.
Depending on the experimental condition, the task may also include a description of the system and a labeled support batch.

In the following sections, we first describe the controllable task axes and define the structure of the \name{} task. 
We then explain how \name{} generates the samples contained in these tasks from physical simulators.
\Cref{fig:data_generation} provides an overview of the \name{} framework with an illustrative example.
\subsection{Controllable task axes}
\setlength{\textfloatsep}{8pt}
Each \name{} task is instantiated by selecting one setting along each of four controllable axes: domain context, observation noise, labeled examples, and submission mode.
\begin{description}[leftmargin=0pt,labelsep=0.5em,itemsep=2pt,topsep=3pt,parsep=0pt,font=\normalfont\bfseries]
    \item[Domain context.]
    This axis controls whether the agent receives domain context, including a detailed description of the system being analyzed, the meanings of the observed time-series channels, and information about the possible ground-truth classes.
    Removing domain context requires the agent to rely exclusively on patterns in the data.
    \item[Observation noise.]
    This axis controls the level of noise added to the time-series observations. 
    The noise level is parameterized by a scalar \(\kappa\); increasing \(\kappa\) adds more noise to the measurements, making root-cause interventions harder to identify.
    \item[Labeled examples.]
    This axis controls whether the task includes a labeled support batch in addition to the unlabeled test batch.
    The support batch contains an equal number of time-series samples from each class, all generated by the same simulator as the test samples and paired with their ground-truth labels.
    This axis tests whether agents can infer diagnostic patterns from the labeled examples and apply them to the test samples.
    \item[Submission mode.]
    The instruction \(\mathcal{S}\) determines the required submission mode.
    In direct-answer mode \(\mathcal{S}_{\mathrm{direct}}\), the agent must directly return a prediction for each time-series observation in the batch.
    In programmatic-submission mode \(\mathcal{S}_{\mathrm{prog}}\), the agent must write a single reusable Python script that takes as input the time-series observations and returns a class prediction for each time-series observation.
    Because the submitted script can be evaluated on observations that were not available during the episode, this mode also enables measurement of out-of-sample generalization.
    This axis helps determine whether LLM agents can translate their decision-making strategies into a reusable program.
\end{description}

\subsection{Root-cause task definition}
A \name{} task consists of three components:
\begin{enumerate}[leftmargin=*,nosep]
    \item The initial prompt containing the task instructions and, depending on the experimental condition, a textual description of the simulated system, the observed channels, and the candidate ground-truth classes;
    \item An unlabeled test batch \(B_{\mathrm{test}} = \{\tilde{\mathbf{y}}_i\}_{i=1}^{n_{\mathrm{test}}}\) containing the time-series samples for which the agent must return predictions.
    \item An optional labeled support batch \(D_{\mathrm{sup}} = \{(\tilde{\mathbf{y}}_j, c_j)\}_{j=1}^{n_{\mathrm{sup}}}\), where each \(c_j\) is the corresponding ground-truth label.
\end{enumerate}
Each \(\tilde{\mathbf{y}}_i \in \mathbb{R}^{T \times C}\) is a multivariate time-series observation with \(T\) time steps and \(C\) observed channels, and constitutes one classification sample. 
Its associated class indicates either that no intervention occurred or that one candidate simulator parameter changed during the corresponding simulation.

At test time, the agent receives one instantiated task and either returns one prediction for each sample in the test batch in \(\mathcal{S}_{\mathrm{direct}}\) mode or submits a single reusable Python script in \(\mathcal{S}_{\mathrm{prog}}\) mode, which is applied independently to every sample in the batch.
From a machine-learning perspective, each task is therefore a batched closed-set classification problem: one class label must be assigned to every sample in the test batch.

We refer to one complete agent interaction with an instantiated task as an evaluation episode.
An episode begins when the agent receives the task prompt and associated files and ends when the agent returns control.

\subsection{Physical simulator interface}
\name{} generates time-series observations from dynamical systems of the form
\[
\mathbf{y} = \operatorname{Sim}(\boldsymbol{\theta}, \mathbf{x}_0),
\qquad
\tilde{\mathbf{y}} \sim p_{\kappa}(\cdot \mid \mathbf{y}).
\]
\(\operatorname{Sim}\) denotes a physical simulator that generates a noise-free trajectory \(\mathbf{y}\) from a parameter vector \(\boldsymbol{\theta}\) and an initial state \(\mathbf{x}_0\).
In the bouncing-ball simulator used as an illustrative example, \(\boldsymbol{\theta}\) comprises the coefficient of restitution, gravitational acceleration, drag coefficient, and mass, while \(\mathbf{x}_0\) comprises the initial velocity and initial height.
The conditional distribution \(p_\kappa(\cdot\mid\mathbf{y})\) defines the observation model from which the noisy time-series observation \(\tilde{\mathbf{y}}\) is sampled.
It is parameterized by \(\kappa\), a scalar factor controlling the intensity of the measurement corruption.

To generate an intervention sample, we draw an initial state and parameter vector, an intervention time \(t_{\mathrm{int}}\), a root-cause parameter \(\theta_i\), and a post-intervention value \(\theta_i'\).
We model the intervention as a single instantaneous step change.
Formally, let \(\boldsymbol{\theta}^{(i,+)}\) denote the parameter vector obtained by replacing only the \(i\)-th component of \(\boldsymbol{\theta}\) with \(\theta_i'\).
\[
    \boldsymbol{\theta}^{(i)}(t) =
    \begin{cases}
        \boldsymbol{\theta}, & t < t_{\mathrm{int}},\\
        \boldsymbol{\theta}^{(i,+)}, & t \geq t_{\mathrm{int}}.
    \end{cases}
\]
We denote the trajectory generated under this time-varying parameter schedule by \((\mathbf{y}^{(i)})\).

To prevent agents from assuming that a parameter always changes, we also generate no-intervention (\(\mathrm{NI}\)) samples.
For these samples, all parameters remain fixed throughout the simulation, producing the corresponding no-intervention time-series observation \(\tilde{\mathbf{y}}^{\mathrm{NI}}\).
Each generated sample therefore belongs either to the no-intervention class or to one intervention class associated with a parameter in \(\theta\).
The sample label identifies the changed parameter for intervention samples and is \(\mathrm{NI}\) otherwise.

\subsection{Solvability filtering}
\label{sec:solvability-filtering}
To reduce the risk of retaining intervention samples whose root cause is ambiguous to infer from the observed trajectory, we apply a post-generation filtering procedure to each noise-free candidate trajectory.
The procedure targets three sources of ambiguity:
(i) the intervention produces no detectable effect;
(ii) the observations do not establish that a parameter changed during the simulation, because the same trajectory can be generated using the post-intervention parameter value throughout the run; and
(iii) an intervention on another candidate parameter can produce an indistinguishable trajectory.

For example, in the BallDrop simulator of \Cref{fig:data_generation}, changing the coefficient of restitution after the final bounce has no effect because restitution influences only subsequent impacts (i).
Changing it before the first bounce can produce the same trajectory as using the post-intervention restitution value from the beginning, because the baseline value never affects an observed impact (ii).
After the ball has settled into static contact, changes in gravity and mass can become observationally confounded because the ground-contact signal depends on the product \(mg\) (iii).
Exhaustively identifying all such ambiguities would require solving, for every candidate trajectory, a global optimization problem over alternative initial conditions, parameter values, intervention targets, and intervention times.
We therefore use a practical heuristic consisting of two generic probe trajectories and simulator-specific rejection rules.
For each noise-free intervention trajectory \(\mathbf{y}^{(i)}\), generated by a known intervention on \(\theta_i\), we construct:
\begin{enumerate}[leftmargin=*,topsep=3pt,nosep]
    \item the no-intervention probe, generated from the same initial state and baseline parameters but without an intervention;
    \item the static-change probe, generated using the post-intervention parameter vector \(\boldsymbol{\theta}^{(i,+)}\) throughout the simulation.
\end{enumerate}
We reject the intervention sample when its trajectory is not sufficiently distinguishable from either probe.
Specifically, in each channel, we divide the root mean squared difference between the intervention trajectory and the probe trajectory by the expected observation noise in that channel.
We consider two trajectories distinguishable when this dimensionless effect-to-noise ratio exceeds the global threshold \(\rho=2\) in at least one observed channel.
The no-intervention and static-change probes address cases (i) and (ii), respectively.
To address case (iii), we apply manually specified, simulator-specific rejection rules derived from the known system dynamics and observed channels.
These rules reject parameter and intervention configurations with known alternative root-cause explanations.
We provide additional details in \Cref{app:solvability-filtering}.
To prevent no-intervention samples from being classified trivially, we retain a no-intervention sample only if at least two distinct intervention root causes with the same initial parameter configuration pass the filtering procedure.

\FloatBarrier
\begin{figure*}[t]
\centering
\includegraphics[width=0.96\textwidth]{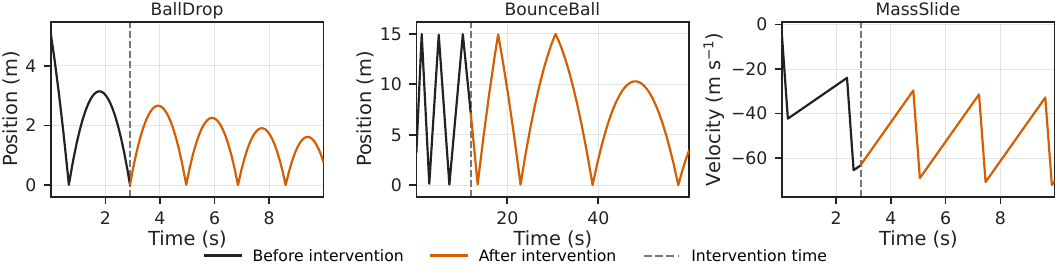}
\Description{Three horizontally arranged time-series plots showing position for BallDrop and BounceBall and velocity for MassSlide. Each plot marks the intervention time with a dashed vertical line.}
\caption{Representative noise-free signals from one observed channel for each simulator.
Black and orange segments show the trajectory before and after the intervention time, respectively, which is marked by a dashed vertical line.
The panels show position for \texttt{BallDrop} and \texttt{BounceBall}, and velocity for \texttt{MassSlide}.
The interventions increase the coefficient of restitution in \texttt{BallDrop} and the inclination angle in \texttt{BounceBall}, and decrease the plane inclination in \texttt{MassSlide}.
Additional examples with observation noise and all observed channels are provided in Appendix~\ref{app:simulator-catalog}.}
\label{fig:simulator-signal-examples}
\end{figure*}

\section{Experimental Setup}
\label{sec:experimental-setup}

\subsection{Physical simulators}
For our experiments, we use the \name{} framework to generate tasks from three simulators of interpretable physical systems:
\begin{itemize}[leftmargin=*,topsep=3pt,nosep]
    \item \texttt{BallDrop}: A ball moving vertically under gravity and bouncing on a ground surface.
    \item \texttt{BounceBall}: A damped mass moving on a tilted track between two rigid walls.
    \item \texttt{MassSlide}: A mass moving along a tilted plane under a periodically applied force.
\end{itemize}
The observation model \(p_{\kappa}\) is simulator-specific and combines multiple noise components, such as Gaussian noise with channel-specific heteroscedastic variance, smooth temporally correlated drift, and transition-localized noise.
For each simulator, we define a reference noise scale, \(\kappa_{\mathrm{ref}}\), that jointly scales these noise components.
We consider two noise regimes: \(\kappa_{\mathrm{low}} = \kappa_{\mathrm{ref}} \qquad \text{and} \qquad \kappa_{\mathrm{high}} = 4\kappa_{\mathrm{ref}}\).

\Cref{fig:simulator-signal-examples} shows representative noise-free signals containing a single parameter intervention for each simulator.
Appendix~\ref{app:simulator-catalog} provides, for each simulator, the simulator description, parameter sampling intervals, candidate root-cause parameters, observed channels, low-noise per-channel signal-to-noise ratios, and representative time-series examples.

\begin{table*}[t]
\centering
\caption{Mean accuracy and exploration metrics per episode for the main experiment, aggregated across simulators.
Accuracy values are reported as the mean $\pm$ within-simulator standard deviation; random-chance accuracy is 0.189.
Python calls indicate the total number of Python interpreter invocations, while Python statements count statements across generated Python source versions.
Plot counts report the number of generated \(\mid\) inspected plots.
Console-output entries report the number of numeric values printed \(\mid\) the estimated percentage of tokens attributable to console output.
See Appendix~\ref{app:metric-computation} for computation details.}
\label{tab:acc_main_experiment_table}
\begingroup
\normalsize
\providecommand{\scorepm}[2]{{\normalsize $#1 \pm #2$}}

\setlength{\tabcolsep}{0pt}
\setlength{\aboverulesep}{0pt}
\setlength{\belowrulesep}{2pt}
\renewcommand{\arraystretch}{1.05}

\begin{tabular*}{\textwidth}{@{\extracolsep{\fill}}lcccccc@{}}
\toprule
& \multicolumn{3}{c}{\texttt{Low Noise}}
& \multicolumn{3}{c}{\texttt{High Noise}} \\
\cmidrule(lr){2-4}\cmidrule(lr){5-7}
& \multicolumn{1}{c}{\(\mathcal{S}_{\mathrm{direct}}\)}
& \multicolumn{2}{c}{\(\mathcal{S}_{\mathrm{prog}}\)}
& \multicolumn{1}{c}{\(\mathcal{S}_{\mathrm{direct}}\)}
& \multicolumn{2}{c}{\(\mathcal{S}_{\mathrm{prog}}\)} \\
\cmidrule(lr){2-2}\cmidrule(lr){3-4}
\cmidrule(lr){5-5}\cmidrule(lr){6-7}
& \texttt{Episode}
& \texttt{Episode}
& \texttt{Held-out}
& \texttt{Episode}
& \texttt{Episode}
& \texttt{Held-out} \\
\midrule
& \multicolumn{6}{c}{\textit{Accuracy}} \\
\texttt{gpt-5.5}
& {\bfseries\boldmath \scorepm{0.933}{0.056}}
& {\bfseries\boldmath \scorepm{0.853}{0.143}}
& {\bfseries\boldmath \scorepm{0.671}{0.096}}
& {\bfseries\boldmath \scorepm{0.833}{0.194}}
& {\bfseries\boldmath \scorepm{0.733}{0.227}}
& {\bfseries\boldmath \scorepm{0.547}{0.138}} \\

\texttt{gemini-3.1-pro}
& \scorepm{0.760}{0.115}
& \scorepm{0.733}{0.153}
& \scorepm{0.604}{0.128}
& \scorepm{0.560}{0.221}
& \scorepm{0.547}{0.169}
& \scorepm{0.440}{0.085} \\

\texttt{claude-opus-4.6}
& \scorepm{0.853}{0.115}
& \scorepm{0.687}{0.122}
& \scorepm{0.614}{0.084}
& \scorepm{0.593}{0.170}
& \scorepm{0.593}{0.190}
& \scorepm{0.499}{0.129} \\

\texttt{minimax-m2.7}
& \scorepm{0.293}{0.134}
& \scorepm{0.280}{0.126}
& \scorepm{0.287}{0.031}
& \scorepm{0.300}{0.165}
& \scorepm{0.353}{0.165}
& \scorepm{0.299}{0.045} \\

\specialrule{\lightrulewidth}{0pt}{6pt}
& \multicolumn{6}{c}{\textit{Python: calls \(\mid\) statements}} \\
\texttt{gpt-5.5} & 12.800 | 326.600 & \multicolumn{2}{c}{16.067 | 638.267} & 13.867 | 377.333 & \multicolumn{2}{c}{17.600 | 641.400} \\
\texttt{gemini-3.1-pro} & 39.000 | 1339.400 & \multicolumn{2}{c}{43.333 | 1404.600} & 44.467 | 1220.800 & \multicolumn{2}{c}{48.333 | 1517.067} \\
\texttt{claude-opus-4.6} & 23.067 | 2120.933 & \multicolumn{2}{c}{29.867 | 1836.667} & 26.467 | 2065.333 & \multicolumn{2}{c}{33.267 | 2467.867} \\
\texttt{minimax-m2.7} & 21.933 | 987.600 & \multicolumn{2}{c}{63.600 | 1766.800} & 20.133 | 1021.067 & \multicolumn{2}{c}{55.000 | 1706.133} \\

\midrule
& \multicolumn{6}{c}{\textit{Plot counts: generated \(\mid\) inspected}} \\
\texttt{gpt-5.5} & 2.067 | 0.400 & \multicolumn{2}{c}{0.000 | 0.000} & 2.333 | 0.733 & \multicolumn{2}{c}{0.000 | 0.000} \\
\texttt{gemini-3.1-pro} & 3.933 | 0.000 & \multicolumn{2}{c}{2.400 | 0.000} & 5.133 | 0.000 & \multicolumn{2}{c}{1.667 | 0.000} \\
\texttt{claude-opus-4.6} & 0.000 | 0.000 & \multicolumn{2}{c}{0.000 | 0.000} & 0.000 | 0.000 & \multicolumn{2}{c}{0.200 | 0.200} \\
\texttt{minimax-m2.7} & 0.000 | -- & \multicolumn{2}{c}{0.000 | --} & 0.000 | -- & \multicolumn{2}{c}{0.000 | --} \\

\midrule
& \multicolumn{6}{c}{\textit{Console output: numeric values printed \(\mid\) token share (\%)}} \\
\texttt{gpt-5.5} & 3447 | 77.094 & \multicolumn{2}{c}{4775 | 77.682} & 4781 | 79.967 & \multicolumn{2}{c}{4737 | 77.722} \\
\texttt{gemini-3.1-pro} & 3024 | 46.984 & \multicolumn{2}{c}{2299 | 45.923} & 3631 | 51.802 & \multicolumn{2}{c}{4796 | 55.350} \\
\texttt{claude-opus-4.6} & 5416 | 39.871 & \multicolumn{2}{c}{5016 | 43.480} & 6630 | 45.256 & \multicolumn{2}{c}{6943 | 41.131} \\
\texttt{minimax-m2.7} & 1674 | 39.052 & \multicolumn{2}{c}{2679 | 29.277} & 1500 | 40.340 & \multicolumn{2}{c}{2546 | 30.067} \\

\bottomrule
\end{tabular*}
\endgroup


%
\end{table*}

\subsection{Metrics}
\label{sec:metrics}
Our evaluation uses two complementary approaches.
On the one hand, we measure predictive performance using accuracy; on the other hand, we extract quantitative metrics from the agentic log trace generated throughout each episode.
For tasks using \(\mathcal{S}_{\mathrm{direct}}\), we measure accuracy only on the episode batch, that is, the test files visible to the agent.
For tasks using \(\mathcal{S}_{\mathrm{prog}}\), we report accuracy both on the episode batch and on a held-out batch of additional samples that is inaccessible to the agent during the episode, thereby measuring generalization.
From the agentic log traces, we extract metrics that characterize overall resource use and exploration behavior.
Specifically, we measure episode token usage and its corresponding cost, episode time, and the number of tool calls.
We additionally report fine-grained exploration and interaction metrics, including the numbers of Python calls and Python statements; the numbers of plots generated and inspected; the number of numeric values exposed through console output; and the cumulative console-output token share.
The latter is defined as the estimated number of console-output tokens processed across all agent iterations, divided by the total number of input-context tokens processed across those iterations.
Simulator-specific tables report the mean and standard deviation across repetition seeds.
Aggregate tables report the pooled mean across simulators and repetition seeds.
To summarize variability without conflating it with simulator-specific difficulty, we first subtract each simulator's mean and then compute the standard deviation of the pooled within-simulator residuals.
Definitions and computation details for all metrics extracted from the agentic log trace are provided in \Cref{app:metric-computation}.

\subsection{Evaluated LLM agents}
Each evaluated configuration is an agent system comprising a language model, a terminal-based agent scaffold, and the associated inference settings.
For readability, we refer to each complete agent-system configuration using the name of its underlying model.
Specifically, we pair \texttt{gpt-5.5}, \texttt{gemini-3.1-pro}, and \texttt{claude-opus-4.6} with Codex, Gemini CLI, and Claude Code, respectively.
We additionally evaluate the open-weight model \texttt{minimax-\allowbreak m2.7} through OpenCode.
For LLM agents with a reasoning-effort setting, we set the reasoning effort to \texttt{high}.
All agents are evaluated under the same submission modes, file layout, and scoring procedure.
Detailed information about the agent profiles, execution platforms, package versions, and reasoning settings is provided in Appendix~\ref{app:agent-platforms-and-profiles}.

\subsection{Runtime environment}
Each evaluation episode is executed in an isolated containerized environment.
The agent has access to a Bash terminal and a Python environment with a set of preinstalled libraries.
The list of available Python packages is provided in \Cref{app:agent-runtime-libraries}.
Internet access is disabled.
To ensure that our analysis reflects agents' abilities rather than infrastructure reliability, we restart episodes that terminate because of infrastructure-related failures, such as API outages, platform interruptions, or container initialization failures.

\subsection{Experimental design}
Each experimental condition is defined by its test set and its settings along four controllable task axes: domain context, observation noise, labeled examples, and submission mode.
To support controlled comparisons when varying a task axis, matched experimental conditions use the same test samples.
This ensures that differences between conditions are attributable to the setting being varied rather than to differences in the underlying test samples.
Test batches are sampled from the available pool of samples that have passed the solvability filtering.

Our main experiment comprises four experimental conditions: low-noise direct answer (\(\kappa_{\mathrm{low}}, \mathcal{S}_{\mathrm{direct}}\)), low-noise programmatic submission (\(\kappa_{\mathrm{low}}, \mathcal{S}_{\mathrm{prog}}\)), high-noise direct answer (\(\kappa_{\mathrm{high}}, \mathcal{S}_{\mathrm{direct}}\)), and high-noise programmatic submission (\(\kappa_{\mathrm{high}}, \mathcal{S}_{\mathrm{prog}}\)).
Each condition includes three labeled support samples per class and a detailed textual description of the system.
For each simulator, condition, and repetition seed, we instantiate one task and evaluate one agent on it, yielding one evaluation episode.
Each task contains 10 test samples.
We use five repetition seeds for each simulator and condition.
This yields 60 evaluation episodes per agent: 15 episodes for each of the four conditions, corresponding to 150 test samples per condition.
We then conduct two orthogonal ablations at \(\kappa_{\mathrm{high}}\) under both the \(\mathcal{S}_{\mathrm{direct}}\) and \(\mathcal{S}_{\mathrm{prog}}\) conditions.
In the first ablation, we remove all domain context, including the process description, column names, and class names.\footnote{Class names are replaced with anonymized labels such as \texttt{label\_0}, \ldots, \texttt{label\_n}.}
This setting provides a purely data-driven baseline.
In the second ablation, we remove the labeled examples while retaining the domain context.
Because of cost constraints, we evaluate these ablations only with \texttt{gpt-5.5} and \texttt{gemini-3.1-pro}.

Example prompts for the different settings are provided in \Cref{app:prompt-example-combinations}.
We present the aggregate analysis in the following sections and report paired statistical comparisons and uncertainty estimates in \Cref{app:statistical-analysis}.
We provide results with standard baselines in \Cref{app:classical-baselines}.

\section{Results}
\begin{figure*}[t]
\centering
\includegraphics[width=\textwidth]{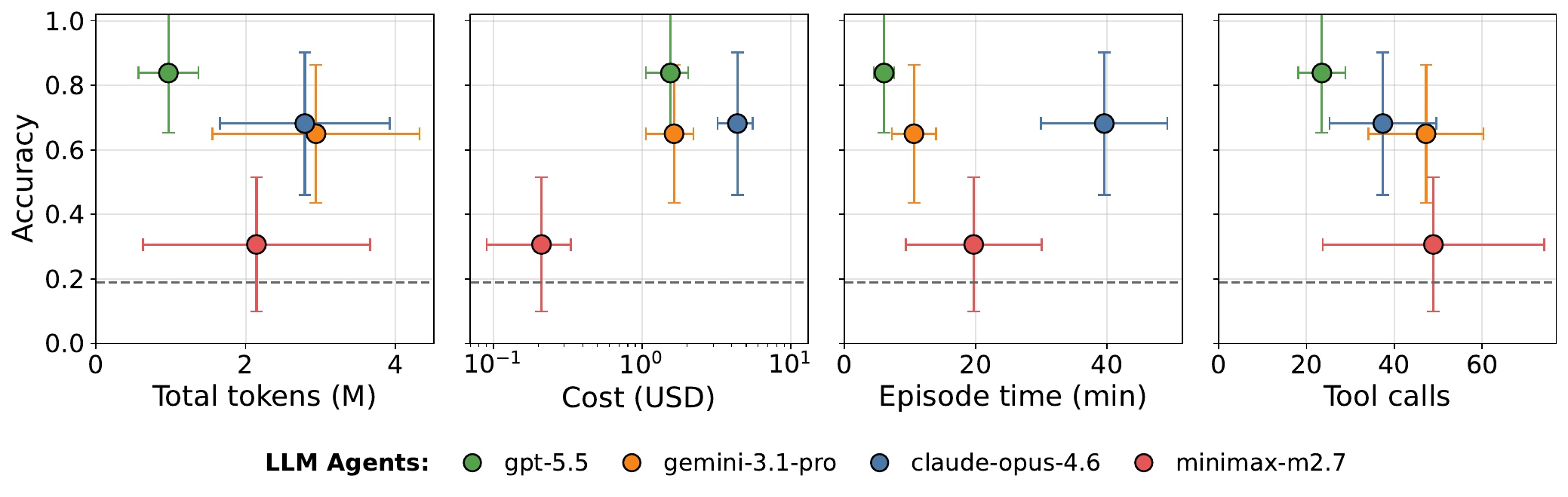}
\Description{Four scatter plots compare model accuracy with token usage, evaluation cost, episode time, and tool calls.
Each point represents one agent-system configuration aggregated over all episodes, with error bars showing population standard deviations.}
\caption{Accuracy as a function of token usage, evaluation cost, episode time, and tool calls.
Each point aggregates all episodes for one agent-system configuration across the four main-experiment conditions, three simulators, and five repetition seeds.}
\label{fig:accuracy-vs-compute}
\end{figure*}

\subsection{Main results}
The first section of Table~\ref{tab:acc_main_experiment_table} reports accuracy averaged over simulators and repetition seeds.
Overall, the \texttt{gpt-5.5} configuration achieves the highest mean accuracy in all six reported accuracy columns.
\texttt{claude-opus-4.6} ranks second in five of the six columns, while \texttt{gemini-3.1-pro} ranks second for low-noise programmatic episode accuracy.
\texttt{minimax-m2.7} performs substantially worse: averaged across the six reported accuracy columns, its accuracy is approximately 46.0 percentage points below that of \texttt{gpt-5.5}.

This large gap between \texttt{gpt-5.5} and \texttt{minimax-m2.7} contrasts with their relatively similar reported performance on SWE-Bench Pro~\citep{minimax_m27_report_2026}, where \texttt{minimax-m2.7} achieves 56.22\% and \texttt{gpt-5.5} achieves 58.6\%.
By contrast, Terminal-Bench~2.0~\citep{merrill2026terminalbench}, which evaluates agents on long-horizon tasks in terminal environments, also shows a substantial performance gap: \texttt{gpt-5.5} achieves 82.2\%, whereas the highest-scoring listed agent using \texttt{minimax-m2.7} achieves 45.1\%, a difference of 37.1 percentage points~\citep{terminalbench_leaderboard}.
Although comparisons across SWE-Bench Pro, Terminal-Bench~2.0, and \name{} are not fully controlled because the tasks and agent configurations differ, the similarly large performance gaps observed on Terminal-Bench~2.0 and \name{} are consistent with the hypothesis that success on \name{} depends not only on code-generation ability, but also on sustained, multi-turn, tool-assisted analysis.

For each of the three closed-weight LLM agents, observed mean episode accuracy decreases as noise increases from \(\kappa_{\mathrm{low}}\) to \(\kappa_{\mathrm{high}}\).
Averaged across the two submission modes, the magnitude of this decrease is 11.0 percentage points for \texttt{gpt-5.5}, 17.7 points for \texttt{claude-opus-4.6}, and 19.3 points for \texttt{gemini-3.1-pro}.
Averaged over noise levels, observed mean episode accuracy is also lower under \(\mathcal{S}_{\mathrm{prog}}\) than under \(\mathcal{S}_{\mathrm{direct}}\) for each configuration.
This pattern suggests that agents struggle under the programmatic interface, which requires compressing their analysis into a reusable per-sample function and removes opportunities for sample-specific judgment.
Within \(\mathcal{S}_{\mathrm{prog}}\), held-out accuracy averaged over noise levels is lower than episode accuracy for all three closed-weight agents, suggesting that the submitted programs do not always capture stable physical signatures of the interventions.

Higher resource use does not correspond monotonically to higher accuracy, as illustrated by \Cref{fig:accuracy-vs-compute}.
The figure compares agent-configuration-level mean accuracy with mean token usage, evaluation cost, episode time, and tool calls across all 60 main-experiment episodes for each configuration.
The \texttt{gpt-5.5} configuration achieves the highest accuracy in every reported condition while using comparatively modest resources, a pattern consistent with more targeted exploration.
The \texttt{claude-opus-4.6} configuration, by contrast, reaches similar accuracy in some conditions but uses substantially more tokens and incurs much higher evaluation cost.
Across conditions, its mean evaluation cost ranges from approximately USD~5.2 to USD~8.6 per episode, compared with USD~1.3 to USD~1.8 for \texttt{gpt-5.5}.
Complete per-simulator cost results are provided in \Cref{app:per-simulator-result-tables}, together with the corresponding accuracy-versus-cost plots in \Cref{app:accuracy-vs-cost}.

\paragraph{Exploration and interaction metrics}
To characterize the exploration strategies used by the evaluated agents, we additionally report in \Cref{tab:acc_main_experiment_table} Python-use, plotting, and console-output metrics.
Several trends emerge.
First, averaged across the three closed-weight configurations and the four main-experiment conditions, more than 4,600 numeric values are exposed through console output per episode.
These results indicate that numerical console output constitutes a substantial part of the evidence made available to the agents during their analyses.
Interestingly, \texttt{minimax-m2.7} exposes substantially fewer numeric values, suggesting less intensive use of numerical console evidence.

Plot inspection is rare among the configurations that support image inspection\footnote{\texttt{minimax-m2.7} does not support image inspection in the evaluated configuration.}.
\texttt{gpt-5.5}, the LLM agent that inspects the most plots, inspects fewer than one plot per episode on average in the direct-answer submission conditions and none in the programmatic-submission conditions.
Notably, \texttt{gemini-3.1-pro} generates 1.7--5.1 plots per episode but does not subsequently inspect them.
This behavior contrasts with conventional exploratory data analysis, in which visualization is commonly used to inspect time-series data.
In summary, these results show that, in contrast to prior work on zero-shot time-series reasoning~\citep{merrill2024language}, closed-weight agents can reason effectively over time-series data in a tool-assisted setting and rely primarily on numerical console output to guide their reasoning and predictions.

The Python-use metrics provide further evidence of differences in exploration behavior and help contextualize the combination of high accuracy and comparatively low token usage observed for \texttt{gpt-5.5}.
Across all four conditions, the \texttt{gpt-5.5} configuration makes substantially fewer Python calls and generates fewer Python statements than the other closed-weight configurations.
\texttt{minimax-m2.7} and \texttt{gemini-3.1-pro} are instead the models that make the most Python calls, while \texttt{claude-opus-4.6} generates the most Python statements.
Console-output shares also show substantial differences across agents.
For \texttt{gpt-5.5}, console output accounts for 77--80\% of its cumulative input context, compared with approximately 29--55\% for the other configurations.

Taken together, these observations suggest that \texttt{gpt-5.5} follows a more compact computational workflow, using Python more selectively while still exposing a substantial amount of numerical evidence through console output.
A complete representative trajectory is provided in \Cref{app:gpt-5-5-ball-drop-direct-example}, and the complete set of trajectories is available in the results section of our project website.

\begin{table*}[t]
\centering
\caption{Accuracy and token usage for the two high-noise ablation conditions, aggregated across simulators.
Entries report the mean \(\pm\) within-simulator standard deviation, followed in parentheses by the signed change relative to the corresponding non-ablated condition with three labeled examples per class and domain context. Accuracy changes are in percentage points; token values and changes are in millions.}
\label{tab:acc_ablation_training_examples}
\begingroup
\normalsize
\providecommand{\scorepm}[2]{{\normalsize $#1 \pm #2$}}
\providecommand{\ablationpositive}[1]{{\normalsize\textcolor{green!50!black}{+#1}}}
\providecommand{\ablationnegative}[1]{{\normalsize\textcolor{red!70!black}{-#1}}}
\providecommand{\ablationneutral}[1]{\textcolor{black}{#1}}
\providecommand{\tokenincrease}[1]{{\normalsize\textcolor{red!70!black}{+#1}}}
\providecommand{\tokendecrease}[1]{{\normalsize\textcolor{green!50!black}{-#1}}}
\providecommand{\tokenneutral}[1]{\textcolor{black}{#1}}
\providecommand{\ablationsectionheading}[1]{\makebox[0pt][c]{\makebox[0pt][c]{\textit{#1}}\hspace*{53.8pt}}}

\setlength{\tabcolsep}{6pt}
\setlength{\aboverulesep}{0pt}
\setlength{\belowrulesep}{2pt}
\renewcommand{\arraystretch}{1.05}

\begin{tabular*}{\textwidth}{@{\extracolsep{\fill}}lccc@{}}
\toprule
& \multicolumn{1}{c}{\(\mathcal{S}_{\mathrm{direct}}\)} & \multicolumn{2}{c}{\(\mathcal{S}_{\mathrm{prog}}\)} \\
\cmidrule(lr){2-2}\cmidrule(lr){3-4}
& \texttt{Episode} & \texttt{Episode} & \texttt{Held-out} \\
\midrule
& \multicolumn{3}{c}{\ablationsectionheading{Accuracy}} \\
\texttt{gpt-5.5} (No labeled examples) & \scorepm{0.887}{0.125}\enspace(\ablationpositive{5.3}) & \scorepm{0.847}{0.116}\enspace(\ablationpositive{11.3}) & \scorepm{0.506}{0.070}\enspace(\ablationnegative{4.1}) \\
\addlinespace[3pt]
\texttt{gemini-3.1-pro} (No labeled examples) & \scorepm{0.647}{0.159}\enspace(\ablationpositive{8.7}) & \scorepm{0.653}{0.176}\enspace(\ablationpositive{10.7}) & \scorepm{0.444}{0.113}\enspace(\ablationpositive{0.4}) \\
\midrule
\texttt{gpt-5.5} (No domain context) & \scorepm{0.513}{0.202}\enspace(\ablationnegative{32.0}) & \scorepm{0.453}{0.209}\enspace(\ablationnegative{28.0}) & \scorepm{0.316}{0.092}\enspace(\ablationnegative{23.2}) \\
\addlinespace[3pt]
\texttt{gemini-3.1-pro} (No domain context) & \scorepm{0.313}{0.135}\enspace(\ablationnegative{24.7}) & \scorepm{0.347}{0.154}\enspace(\ablationnegative{20.0}) & \scorepm{0.328}{0.043}\enspace(\ablationnegative{11.3}) \\

\specialrule{\lightrulewidth}{0pt}{6pt}
& \multicolumn{3}{c}{\ablationsectionheading{Total tokens (M)}} \\
\texttt{gpt-5.5} (No labeled examples) & \scorepm{0.796}{0.197}\enspace(\tokendecrease{0.177}) & \multicolumn{2}{c}{\scorepm{0.892}{0.289}\enspace(\tokendecrease{0.297})} \\
\addlinespace[3pt]
\texttt{gemini-3.1-pro} (No labeled examples) & \scorepm{1.832}{0.788}\enspace(\tokendecrease{0.964}) & \multicolumn{2}{c}{\scorepm{2.838}{0.980}\enspace(\tokendecrease{0.849})} \\
\midrule
\texttt{gpt-5.5} (No domain context) & \scorepm{1.796}{0.510}\enspace(\tokenincrease{0.824}) & \multicolumn{2}{c}{\scorepm{2.219}{0.773}\enspace(\tokenincrease{1.031})} \\
\addlinespace[3pt]
\texttt{gemini-3.1-pro} (No domain context) & \scorepm{2.506}{1.059}\enspace(\tokendecrease{0.290}) & \multicolumn{2}{c}{\scorepm{4.141}{1.396}\enspace(\tokenincrease{0.454})} \\
\bottomrule
\end{tabular*}
\endgroup


\end{table*}

\subsection{Ablations: removing labeled examples or domain context}
We next examine two ablations: removing all labeled examples while retaining the domain context, and removing the domain context while retaining all labeled examples.
\Cref{tab:acc_ablation_training_examples} reports the results of both ablations, along with their differences from the corresponding non-ablated high-noise conditions, for both accuracy and token consumption.
Removing the domain context reduces average accuracy by 20.0--32.0 percentage points across both LLM agents and both submission modes.
By contrast, removing the labeled time-series examples increases average accuracy by 4.2 percentage points for \texttt{gpt-5.5} and by 6.6 percentage points for \texttt{gemini-3.1-pro}.
Although removing the domain context shortens the input prompt, it increases total token usage in three of the four cases, suggesting that agents may compensate for the missing context with more extensive analysis.
Removing the labeled examples, by contrast, consistently reduces token usage.
In all cases, removing the examples results in a substantial cost reduction.

These results suggest that LLM agents make effective use of domain context and that richer domain context may reduce token consumption, thereby making requests more efficient.
In contrast, labeled time-series examples may be detrimental when the domain context is already well specified, as they increase costs without substantially improving performance.

\section{Conclusion}
In this paper, we introduced \name{}, a framework for generating controlled time-series root-cause attribution tasks.
These tasks evaluate how LLM agents analyze time series generated by dynamical systems, with or without accompanying textual descriptions of those systems.
Our experiments yield several findings.

First, LLM agents explore the data by inspecting numerical console outputs and reasoning over them to guide subsequent analysis and final predictions.
Among the evaluated agent configurations, \texttt{gpt-5.5} achieves the highest accuracy across the main experimental conditions while using comparatively fewer computational resources.
Second, performance generally decreases as observation noise increases and when agents are required to translate their analyses into reusable programs.
Held-out results further indicate that submitted scripts have limited out-of-sample generalization.
Finally, our ablation results show that agents can make effective use of domain context, whereas labeled support sets may increase costs with no reliable evidence of performance improvements.

\paragraph{Limitations.}
First, \name{} uses a closed-ended label set to enable unambiguous evaluation and clean comparisons across conditions.
However, this format does not fully capture the open-ended nature of root-cause analysis, in which analysts must often formulate hypotheses and communicate conclusions without predefined options.
Second, due to the high cost of agent evaluations, we used a small number of repetitions per experimental condition, which limits statistical power.
Third, each model is evaluated using a different agent scaffold.
Hence, observed differences cannot be attributed solely to the underlying LLMs and may also reflect scaffold-specific effects.

\paragraph{Representativeness and intended scope.}
\name{} is a controlled diagnostic testbed rather than a realistic replica of real-world root-cause analysis.
It captures key structural properties of diagnostic time-series tasks, including multivariate observations, unknown-time interventions, nonlinear and event-driven dynamics, delayed and channel-dependent effects, and several forms of measurement noise.
However, it is limited to regularly sampled, fully observed trajectories from low-dimensional mechanical systems with at most one instantaneous parameter change.
Results should therefore be interpreted as evidence of controlled temporal and mechanistic reasoning, not of readiness for complex industrial deployment.
\paragraph{Data and code availability.}
\begin{sloppypar}
We release the experimental code at \href{https://github.com/TommasoBendinelli/TraceBench}{\texttt{github.com/TommasoBendinelli/TraceBench}}, the dataset at \href{https://huggingface.co/datasets/eth-siplab/tracebench}{\texttt{huggingface.co/datasets/eth-siplab/tracebench}}, and the project website, which includes a public leaderboard, at \href{https://tracebench.github.io}{\texttt{tracebench.github.io}}.
\end{sloppypar}
\paragraph{Acknowledgments}
We thank Dominik Hollidt and Demirel Berken for helpful discussions and feedback.
This work was partially supported by RIS Zentralschweiz and NVIDIA Corporation.

\FloatBarrier
\setcounter{dbltopnumber}{\value{TraceBenchSavedDblTopNumber}}

\clearpage
\bibliographystyle{ACM-Reference-Format}
\bibliography{tex/support/references}

\clearpage
\appendix
\crefname{appendix}{Appendix}{Appendices}
\Crefname{appendix}{Appendix}{Appendices}
\crefalias{section}{appendix}
\etocdepthtag.toc{tracebench-appendix}
\def\tracebenchCombinedAppendixMode{1}
\urlstyle{same}

\providecommand{\code}[1]{%
  \begingroup
  \def\UrlFont{\ttfamily}%
  \nolinkurl{#1}%
  \endgroup
}
\providecommand{\titlecode}[1]{\code{#1}}
\providecommand{\termSRD}{\code{SRD}\xspace}
\providecommand{\termMinSRDDistance}{\code{min_srd_distance}\xspace}
\providecommand{\termMinAbsDist}{\code{min_abs_dist}\xspace}

\DefineVerbatimEnvironment{DocCode}{Verbatim}{
  fontsize=\small,
  formatcom=\TraceBenchQuietVerbatimSpacing,
  breaklines=true,
  breakanywhere=true,
  breaksymbolleft={},
  breaksymbolright={}
}
\newcolumntype{P}[1]{>{\raggedright\arraybackslash}p{#1}}

\newenvironment{EnvironmentConfigTableNoMod}{
  \begingroup
  \normalsize
  \renewcommand{\arraystretch}{1.12}
  \setlength{\tabcolsep}{5pt}
  \begin{table*}[p]
  \centering
  \caption{Initial states and parameters.}
  \begin{tabular}{|P{0.35\textwidth}|P{0.27\textwidth}|>{\normalsize}P{0.24\textwidth}|}
  \hline
  \textbf{Agent-facing parameter} & \textbf{Internal naming} & \textbf{Sampling details} \\
  \hline
}{
  \hline
  \end{tabular}
  \end{table*}
  \endgroup
}

\newenvironment{MappingTable}[1]{
  \normalsize
  \begin{table*}[p]
  \centering
  \caption{#1}
  \begin{tabular}{|P{0.37\textwidth}|P{0.33\textwidth}|>{\normalsize}P{0.16\textwidth}|}
  \hline
  \textbf{Agent-facing} & \textbf{Internal naming} & \textbf{Average SNR (low, dB)} \\
  \hline
}{
  \hline
  \end{tabular}
  \end{table*}
}

\newenvironment{SimulationPropertiesTable}[1]{
  \begingroup
  \normalsize
  \let\ident\code
  \begin{table*}[p]
  \centering
  \caption{#1}
  \begin{tabular}{|P{0.43\textwidth}|>{\normalsize}P{0.41\textwidth}|}
  \hline
  \textbf{Property} & \textbf{Example baseline value} \\
  \hline
}{
  \hline
  \end{tabular}
  \end{table*}
  \endgroup
}

\begingroup
\etocstandardlines
\etocsettagdepth{tracebench-main}{none}
\etocsettagdepth{tracebench-appendix}{section}
\etocsetnexttocdepth{section}
\etocsettocstyle{\pdfbookmark[1]{Appendix Contents}{appendix-contents}\section*{Appendix Contents}}{}
\tableofcontents
\endgroup
\clearpage

\phantomsection\label{docstart:kdd_appendix}
\section{Simulator Catalog}
\label{app:simulator-catalog}
\raggedbottom

\bigskip
\noindent\begin{minipage}{\linewidth}
\noindent{\LARGE\bfseries\texttt{BallDrop}\par}
\noindent\textbf{Simulator page}\newline
\noindent\url{https://tracebench.github.io/environments/BallDrop/}\par

\par\smallskip
\noindent\includegraphics[width=\linewidth]{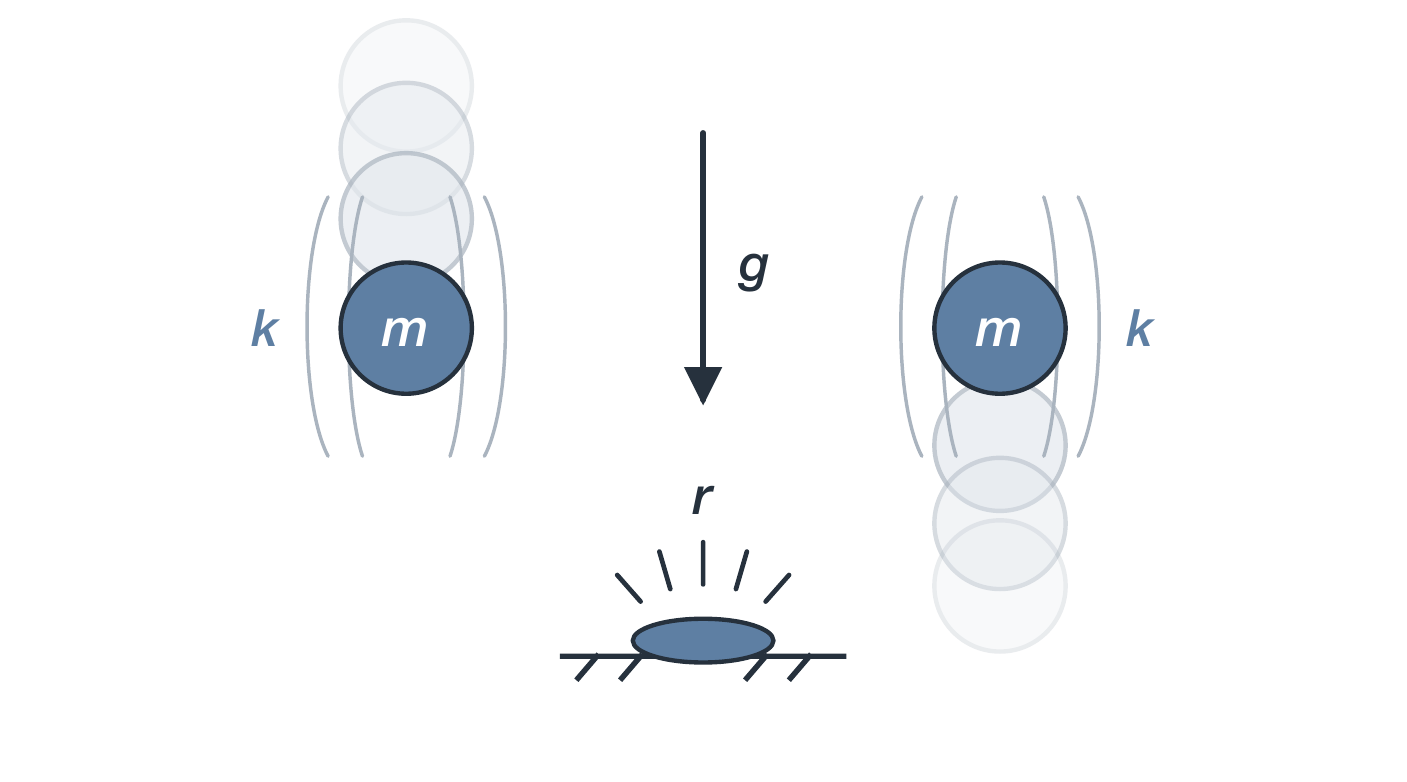}
\Description{Stylized schematic of the BallDrop simulator showing a point-mass ball falling and rebounding under gravity, quadratic air drag, and ground-contact restitution.}
\end{minipage}
\par

\phantomsection\label{app:simulator:balldrop}
\medskip
\noindent{\large\bfseries System description\par}
\begin{DocCode}
These time series were generated by simulating a 1D vertical point-mass ball model under gravity.
While the ball is above the ground, its motion is governed by gravity and quadratic air drag acting opposite the direction of travel, and the position evolves according to the current velocity.
Ground interaction is modeled with a restitution-based hard-stop law.
When the ball reaches the ground with impact speed equal or above a prescribed impact-speed cutoff 0.5 m/s, the impact is treated as instantaneous and the post-impact speed is set by the coefficient of restitution e in [0,1].
The rebound velocity points upward and its magnitude equals e times the pre-impact speed.
When the impact speed is below the threshold, the ball does not rebound and instead enters static contact at the ground.

\end{DocCode}

\noindent\textbf{Observed channels:} position, velocity, and ground-contact impulse.
\par\smallskip
\noindent\textbf{Candidate labels:} \emph{coefficient of restitution},
\emph{mass}, \emph{drag coefficient}, \emph{gravity acceleration}, and
\emph{no parameter change}.

\begin{table}[H]
\centering
\caption{BallDrop simulation settings used by the released trajectories.}
\label{tab:balldrop-release-simulation}
\begingroup
\normalsize
\setlength{\tabcolsep}{0pt}
\setlength{\aboverulesep}{0pt}
\setlength{\belowrulesep}{2pt}
\renewcommand{\arraystretch}{1.05}
\begin{tabular*}{\columnwidth}{@{\extracolsep{\fill}}lr@{}}
\toprule
Simulation End & 10 s \\
Sampling Frequency & 400 Hz \\
Number of Rows & 3999 \\
\bottomrule
\end{tabular*}
\endgroup

\end{table}
\begin{table}[H]
\centering
\caption{Simulator-specific rejection rules for BallDrop.}
\label{tab:balldrop-manual-rejection-rules}
\begingroup\small 
\setlength{\tabcolsep}{4pt} 
\setlength{\aboverulesep}{0pt} 
\setlength{\belowrulesep}{2pt} 
\renewcommand{\arraystretch}{1.08} 
\begin{tabularx}{\columnwidth}{@{}>{\raggedright\arraybackslash}p{0.26\columnwidth}>{\raggedright\arraybackslash}X@{}}
\toprule 
Applies to & Rule \\ 
\midrule 
All interventions & The first apex after the first detected divergence must exceed \(0.1\,\mathrm{m}\) in at least one of the intervention and baseline traces, and each trace must contain fewer than 25 qualifying rebounds. \\
\addlinespace[2pt] 
Qualifying rebound & The ground-contact signal must exceed \(1\), the incoming speed must exceed \(0.5\,\mathrm{m\,s^{-1}}\), and the rebound speed must exceed \(0.1\,\mathrm{m\,s^{-1}}\); events separated by less than \(0.03\,\mathrm{s}\) are merged. \\
\addlinespace[2pt] 
Drag, mass, and restitution & A qualifying rebound must occur both before and after the first detected divergence. \\
\addlinespace[2pt] 
Drag & The normalized-DTW velocity distance over the following \(0.2\,\mathrm{s}\) must exceed \(0.1\), or the terminal-velocity rule must hold. \\
\addlinespace[2pt] 
Terminal-velocity rule & Accept a near-terminal descent after a rebound or one that ends above \(0.1\,\mathrm{m}\) before the trace ends. \\
\bottomrule 
\end{tabularx}
\endgroup 

\end{table}

\begin{table}[H]
\centering
\caption{Empirical BallDrop retained-pool variable summaries.}
\label{tab:balldrop-release-retained-variables}
\begingroup\normalsize
\setlength{\tabcolsep}{0pt}
\renewcommand{\arraystretch}{1.05}
\begin{tabular*}{\columnwidth}{@{\extracolsep{\fill}}lrrr@{}}
\toprule
variable & min & median & max \\
\midrule
\texttt{initial\_height} & 0.19122 & 3.0483 & 5.769 \\
\texttt{initial\_velocity} & -9.7276 & -4.8861 & 1.1469 \\
\texttt{mass} & 0.5 & 3.2 & 19.434 \\
\texttt{drag\_coeff} & 0 & 0.5 & 6 \\
\texttt{restitution} & 0.377 & 0.93 & 0.99 \\
\texttt{gravity} & 2 & 5 & 10 \\
\bottomrule
\end{tabular*}
\par\smallskip\noindent
\begin{minipage}{0.98\columnwidth}
\footnotesize Empirical summaries are sample-weighted across every retained released sample.
\end{minipage}
\endgroup

\end{table}
\FloatBarrier
\begin{table}[H]
\centering
\caption{Exact BallDrop observation-noise coefficients.}
\label{tab:balldrop-release-noise}
\begingroup\normalsize
\setlength{\tabcolsep}{0pt}
\renewcommand{\arraystretch}{1.05}
\begin{tabular*}{\columnwidth}{@{\extracolsep{\fill}}lr@{}}
\toprule
Coefficient & \(\kappa_{\mathrm{ref}}\) \\
\midrule
\texttt{force\_base\_sigma\_scale} & 0.002 \\
\texttt{force\_event\_sigma\_scale} & 0.002 \\
\texttt{force\_hetero\_sigma\_scale} & 0.002 \\
\texttt{position\_base\_sigma\_scale} & 0.002 \\
\texttt{position\_drift\_sigma\_scale} & 0.002 \\
\texttt{position\_event\_sigma\_scale} & 0.002 \\
\texttt{position\_quant\_step\_floor} & 0.0001 \\
\texttt{position\_quant\_step\_scale} & 0.002 \\
\texttt{velocity\_base\_sigma\_scale} & 0.002 \\
\texttt{velocity\_drift\_sigma\_scale} & 0.002 \\
\texttt{velocity\_event\_sigma\_scale} & 0.002 \\
\texttt{velocity\_hetero\_sigma\_scale} & 0.002 \\
\bottomrule
\end{tabular*}
\endgroup

\end{table}

\begin{figure*}[t]
\centering
\begin{minipage}[t]{0.49\textwidth}
\centering
\includegraphics[width=\linewidth]{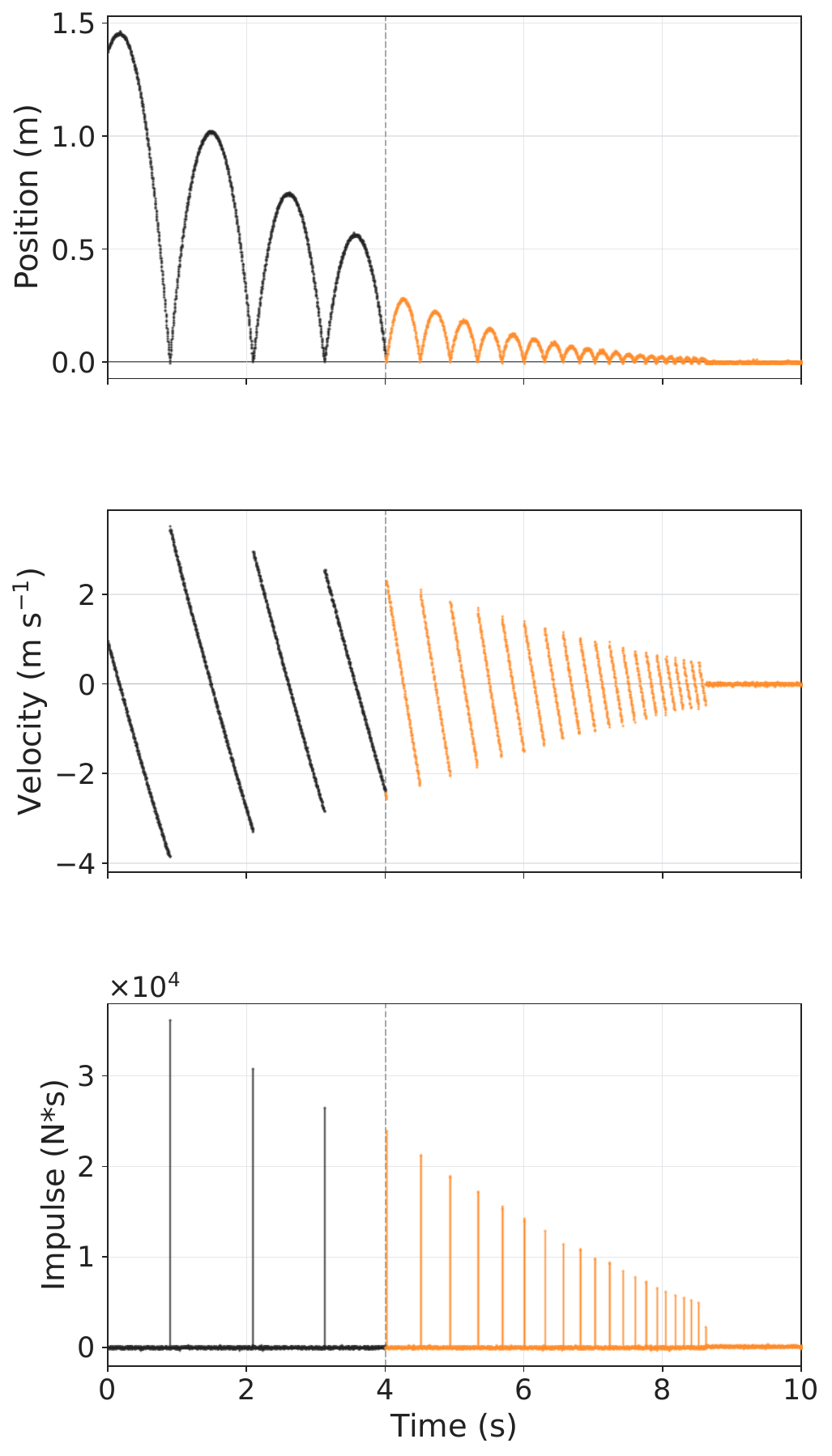}
\par\smallskip
\textbf{(a) Low observation noise}
\end{minipage}\hfill
\begin{minipage}[t]{0.49\textwidth}
\centering
\includegraphics[width=\linewidth]{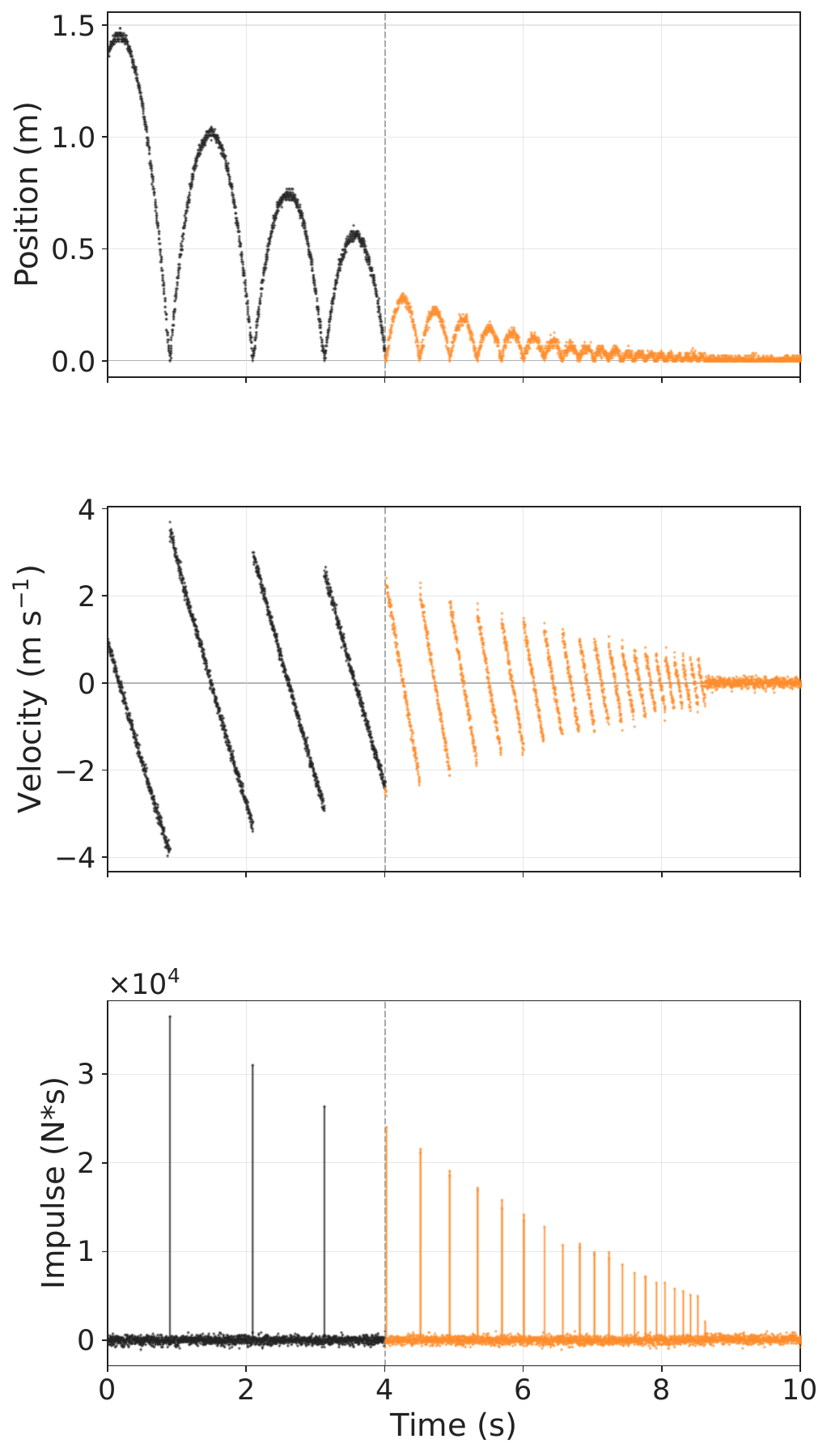}
\par\smallskip
\textbf{(b) High observation noise}
\end{minipage}
\Description{Two side-by-side sets of three aligned BallDrop time-series panels showing position, velocity, and ground-contact impulse (N*s) before and after a gravity intervention. The left set uses low observation noise and the right set uses high observation noise.}
\caption{Position, velocity, and ground-contact impulse (N*s) channels from BallDrop under (a) low and (b) high observation noise.}
\label{fig:balldrop_env_noise_comparison}
\label{fig:balldrop_env_low_noise}
\label{fig:balldrop_env_high_noise}
\end{figure*}

\FloatBarrier
\begin{table*}[!htbp]
\centering
\caption{BallDrop released-trajectory SNR in dB as the mean and sample standard deviation.}
\label{tab:balldrop-release-snr}
\begingroup
\normalsize
\providecommand{\scorepm}[2]{{\normalsize $#1 \pm #2$}}
\setlength{\tabcolsep}{0pt}
\setlength{\aboverulesep}{0pt}
\setlength{\belowrulesep}{2pt}
\renewcommand{\arraystretch}{1.05}
\begin{tabular*}{\textwidth}{@{\extracolsep{\fill}}lrr@{}}
\toprule
\texttt{Channel} & \multicolumn{1}{c}{\textbf{Low Noise}} & \multicolumn{1}{c}{\textbf{High Noise}} \\
\midrule
\texttt{Impulse} & \scorepm{21.07}{2.55} & \scorepm{9.03}{2.55} \\
\texttt{Position} & \scorepm{42.41}{3.19} & \scorepm{30.58}{3.14} \\
\texttt{Velocity} & \scorepm{36.58}{2.60} & \scorepm{24.54}{2.60} \\
\bottomrule
\end{tabular*}

\endgroup

\end{table*}
\FloatBarrier

\FloatBarrier
\clearpage

\par\bigskip
\noindent\begin{minipage}{\linewidth}
\noindent{\LARGE\bfseries\texttt{BounceBall}\par}
\noindent\textbf{Simulator page}\par
\noindent\url{https://tracebench.github.io/environments/BounceBall/}\par

\par\smallskip
\noindent\includegraphics[width=\linewidth]{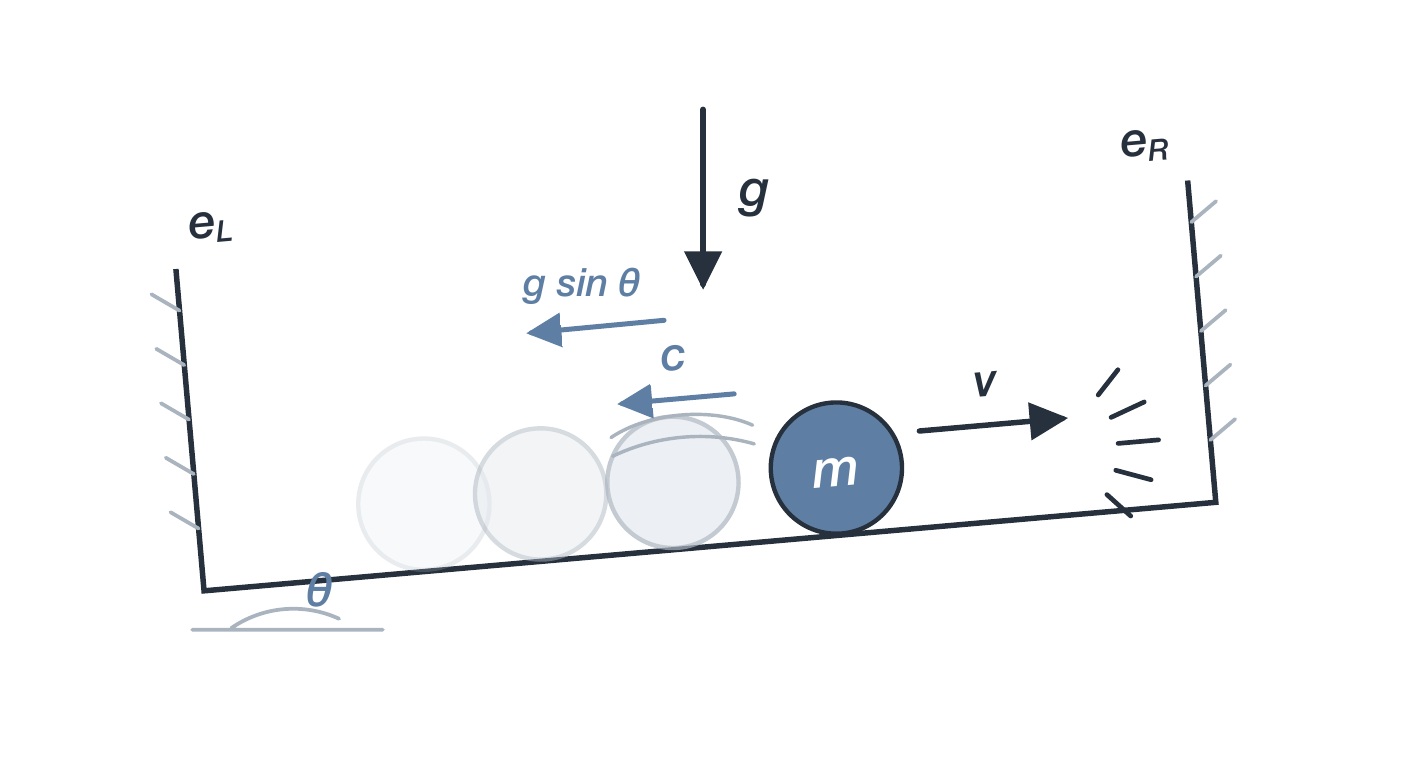}
\Description{Stylized schematic of the BounceBall simulator showing a damped point mass moving along an inclined rail between walls with separate restitution coefficients.}
\end{minipage}
\par

\medskip
\noindent{\large\bfseries System description\par}
\begin{DocCode}
These time series were generated by simulating a damped mass moving along a one-dimensional rail between two rigid walls located at positions x_L and x_R.
The rail may be tilted by a fixed inclination angle.
When the mass is not in contact with either wall, its motion along the rail is governed by viscous drag, proportional to velocity and opposite the direction of motion, and, when the inclination angle is nonzero, by the component of gravity along the rail.
When the mass hits a wall, the collision is modeled as an instantaneous impact: the direction of motion reverses and the post-impact speed is reduced according to that walls' restitution coefficient.

\end{DocCode}

\noindent\textbf{Observed channels:} position, velocity, and right-wall contact force.
\par\smallskip
\noindent\textbf{Candidate labels:} \emph{inclination angle}, \emph{mass},
\emph{left restitution}, \emph{right restitution}, \emph{viscous damping},
and \emph{no parameter change}.

\begin{table}[!htbp]
\centering
\caption{BounceBall simulation settings used by the released trajectories.}
\label{tab:bounceball-release-simulation}
\begingroup
\normalsize
\setlength{\tabcolsep}{0pt}
\setlength{\aboverulesep}{0pt}
\setlength{\belowrulesep}{2pt}
\renewcommand{\arraystretch}{1.05}
\begin{tabular*}{\columnwidth}{@{\extracolsep{\fill}}lr@{}}
\toprule
Simulation End & 60 s \\
Sampling Frequency & 50 Hz \\
Number of Rows & 2999 \\
\bottomrule
\end{tabular*}
\endgroup

\end{table}
\FloatBarrier
\begin{table}[H]
\centering
\caption{Simulator-specific rejection rules for BounceBall.}
\label{tab:bounceball-manual-rejection-rules}
\def\TraceBenchBounceBallImpactRule{Impacts are counted from velocity reversals whose incoming speed exceeds \(0.1\,\mathrm{m\,s^{-1}}\); when velocity reversals cannot be computed, samples with absolute right-contact force above \(10^{-9}\,\mathrm{N}\) are used.}
\def\TraceBenchBounceBallForceRule{In the intervention trace, a qualifying velocity reversal must occur both before and after the intervention time, and right-contact force must exceed \(500\,\mathrm{N}\) on both sides of that time.}
\begingroup\small 
\setlength{\tabcolsep}{4pt} 
\setlength{\aboverulesep}{0pt} 
\setlength{\belowrulesep}{2pt} 
\renewcommand{\arraystretch}{1.08} 
\begin{tabularx}{\columnwidth}{@{}>{\raggedright\arraybackslash}p{0.26\columnwidth}>{\raggedright\arraybackslash}X@{}}
\toprule 
Applies to & Rule \\ 
\midrule 
All interventions & Each usable intervention and baseline trace must contain fewer than 20 impacts. \TraceBenchBounceBallImpactRule \\
\addlinespace[2pt] 
Mass and viscous damping & \TraceBenchBounceBallForceRule \\ 
\bottomrule 
\end{tabularx}
\endgroup 

\end{table}

\begin{table}[H]
\centering
\caption{Empirical BounceBall retained-pool variable summaries.}
\label{tab:bounceball-release-retained-variables}
\begingroup\normalsize
\setlength{\tabcolsep}{0pt}
\renewcommand{\arraystretch}{1.05}
\begin{tabular*}{\columnwidth}{@{\extracolsep{\fill}}lrrr@{}}
\toprule
variable & min & median & max \\
\midrule
\texttt{initial\_position} & 3 & 6.2 & 9.3 \\
\texttt{initial\_velocity} & -17.01 & -8.2 & 10 \\
\texttt{mass} & 0.29498 & 15.435 & 99.287 \\
\texttt{viscous\_damping} & 0.001 & 0.055 & 1.32 \\
\texttt{restitution\_right} & 0.44813 & 0.74813 & 0.9999 \\
\texttt{restitution\_left} & 0.25 & 0.79684 & 0.9999 \\
\texttt{inclination\_angle} & -0.67187 & 0 & 0.95367 \\
\bottomrule
\end{tabular*}
\par\smallskip\noindent
\begin{minipage}{0.98\columnwidth}
\footnotesize Empirical summaries are sample-weighted across every retained released sample.
\end{minipage}
\endgroup

\end{table}
\FloatBarrier
\begin{table}[H]
\centering
\caption{Exact BounceBall observation-noise coefficients.}
\label{tab:bounceball-release-noise}
\begingroup\normalsize
\setlength{\tabcolsep}{0pt}
\renewcommand{\arraystretch}{1.05}
\begin{tabular*}{\columnwidth}{@{\extracolsep{\fill}}lr@{}}
\toprule
Coefficient & \(\kappa_{\mathrm{ref}}\) \\
\midrule
\texttt{position\_base\_sigma\_scale} & 0.0175 \\
\texttt{position\_drift\_sigma\_scale} & 0.004 \\
\texttt{position\_event\_sigma\_scale} & 0.04 \\
\texttt{position\_quant\_step\_floor} & 0.0005 \\
\texttt{position\_quant\_step\_scale} & 0.0025 \\
\texttt{right\_force\_base\_sigma\_scale} & 0.0001 \\
\texttt{right\_force\_event\_sigma\_scale} & 0.004 \\
\texttt{speed\_base\_sigma\_scale} & 0.00088 \\
\texttt{speed\_drift\_sigma\_scale} & 0.0001056 \\
\texttt{speed\_event\_sigma\_scale} & 0.00704 \\
\texttt{speed\_hetero\_sigma\_scale} & 0.0022 \\
\texttt{speed\_post\_event\_sigma\_scale} & 0.00308 \\
\bottomrule
\end{tabular*}
\endgroup

\end{table}

\begin{figure*}[t]
\centering
\begin{minipage}[t]{0.49\textwidth}
\centering
\includegraphics[width=\linewidth]{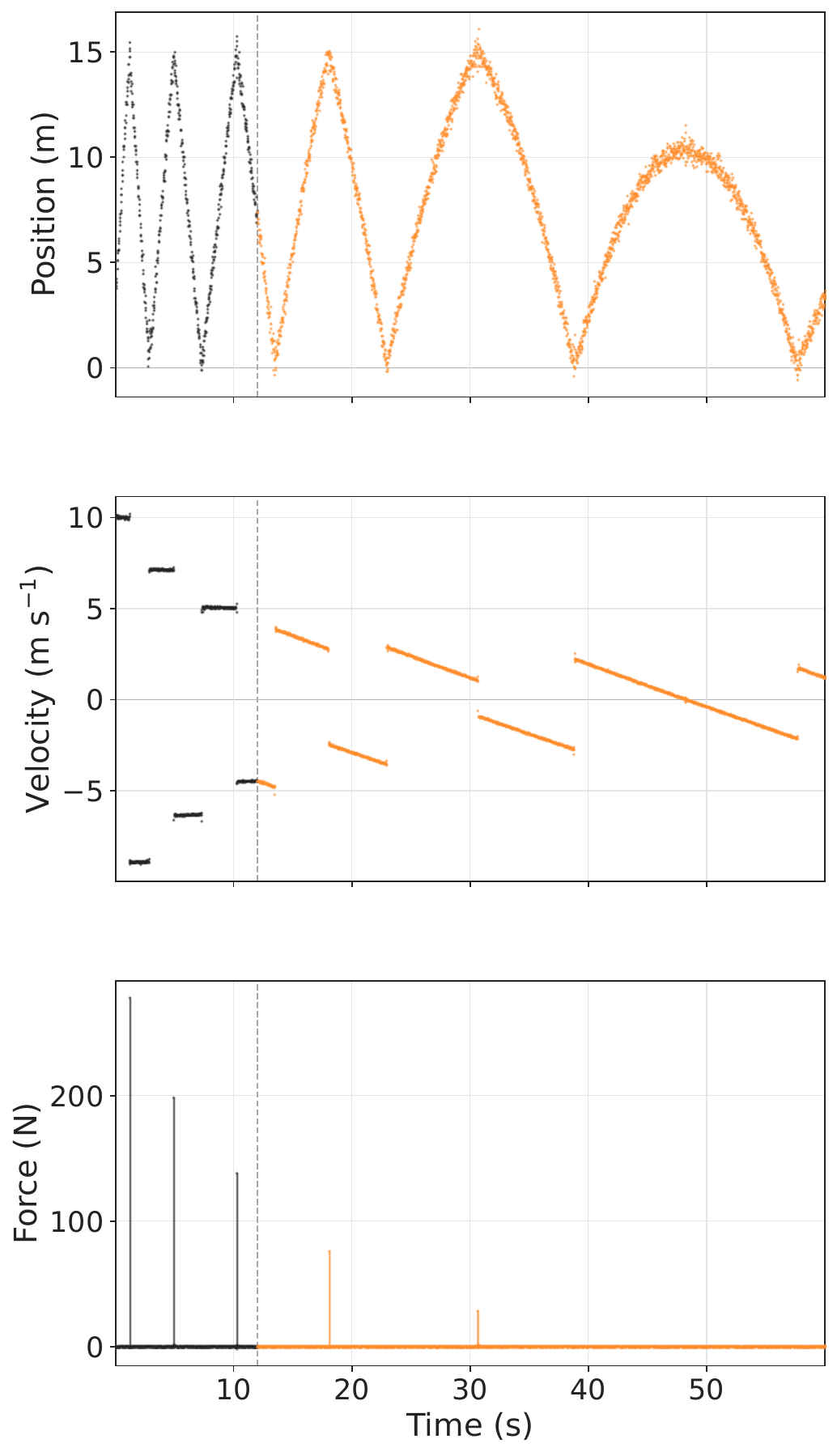}
\par\smallskip
\textbf{(a) Low observation noise}
\end{minipage}\hfill
\begin{minipage}[t]{0.49\textwidth}
\centering
\includegraphics[width=\linewidth]{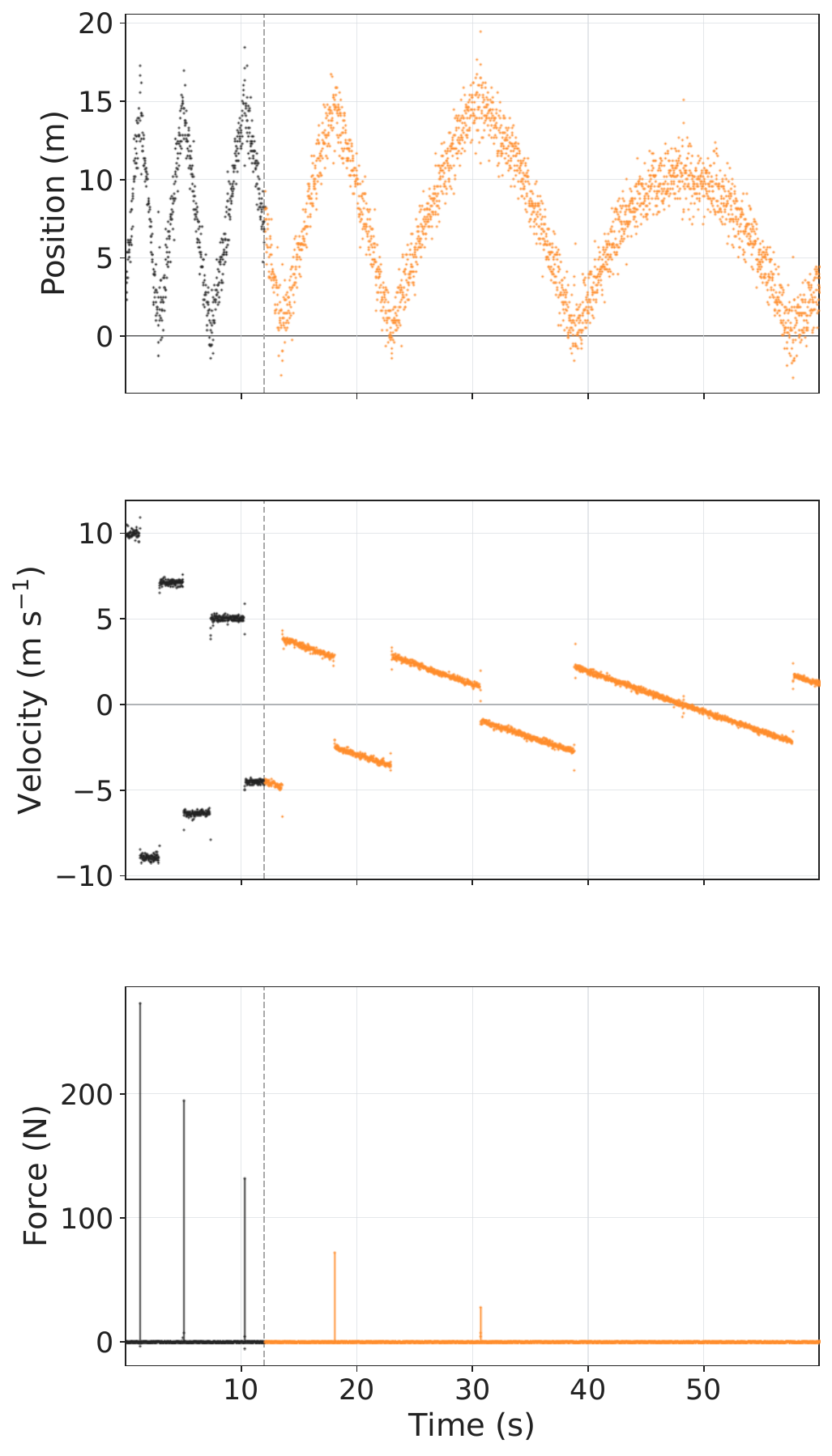}
\par\smallskip
\textbf{(b) High observation noise}
\end{minipage}
\Description{Two side-by-side sets of three aligned BounceBall time-series panels showing position, velocity, and right-wall force before and after an inclination-angle intervention. The left set uses low observation noise and the right set uses high observation noise.}
\caption{Position, velocity, and right-wall force channels from BounceBall under (a) low and (b) high observation noise.}
\label{fig:bounceball_env_noise_comparison}
\label{fig:bounceball_env_low_noise}
\label{fig:bounceball_env_high_noise}
\end{figure*}

\FloatBarrier
\begin{table*}[!htbp]
\centering
\caption{BounceBall released-trajectory SNR in dB as the mean and sample standard deviation.}
\label{tab:bounceball-release-snr}
\begingroup
\normalsize
\providecommand{\scorepm}[2]{{\normalsize $#1 \pm #2$}}
\setlength{\tabcolsep}{0pt}
\setlength{\aboverulesep}{0pt}
\setlength{\belowrulesep}{2pt}
\renewcommand{\arraystretch}{1.05}
\begin{tabular*}{\textwidth}{@{\extracolsep{\fill}}lrr@{}}
\toprule
\texttt{Channel} & \multicolumn{1}{c}{\textbf{Low Noise}} & \multicolumn{1}{c}{\textbf{High Noise}} \\
\midrule
\texttt{Position} & \scorepm{30.02}{1.91} & \scorepm{17.97}{1.91} \\
\texttt{Velocity} & \scorepm{40.98}{2.46} & \scorepm{28.94}{2.46} \\
\texttt{Force} & \scorepm{39.09}{2.45} & \scorepm{27.05}{2.45} \\
\bottomrule
\end{tabular*}

\endgroup

\end{table*}
\FloatBarrier

\clearpage

\par\bigskip
\noindent\begin{minipage}{\linewidth}
\noindent{\LARGE\bfseries\texttt{MassSlide}\par}
\noindent\textbf{Simulator page}\par
\noindent\url{https://tracebench.github.io/environments/MassSlide/}\par

\par\smallskip
\noindent\includegraphics[width=\linewidth]{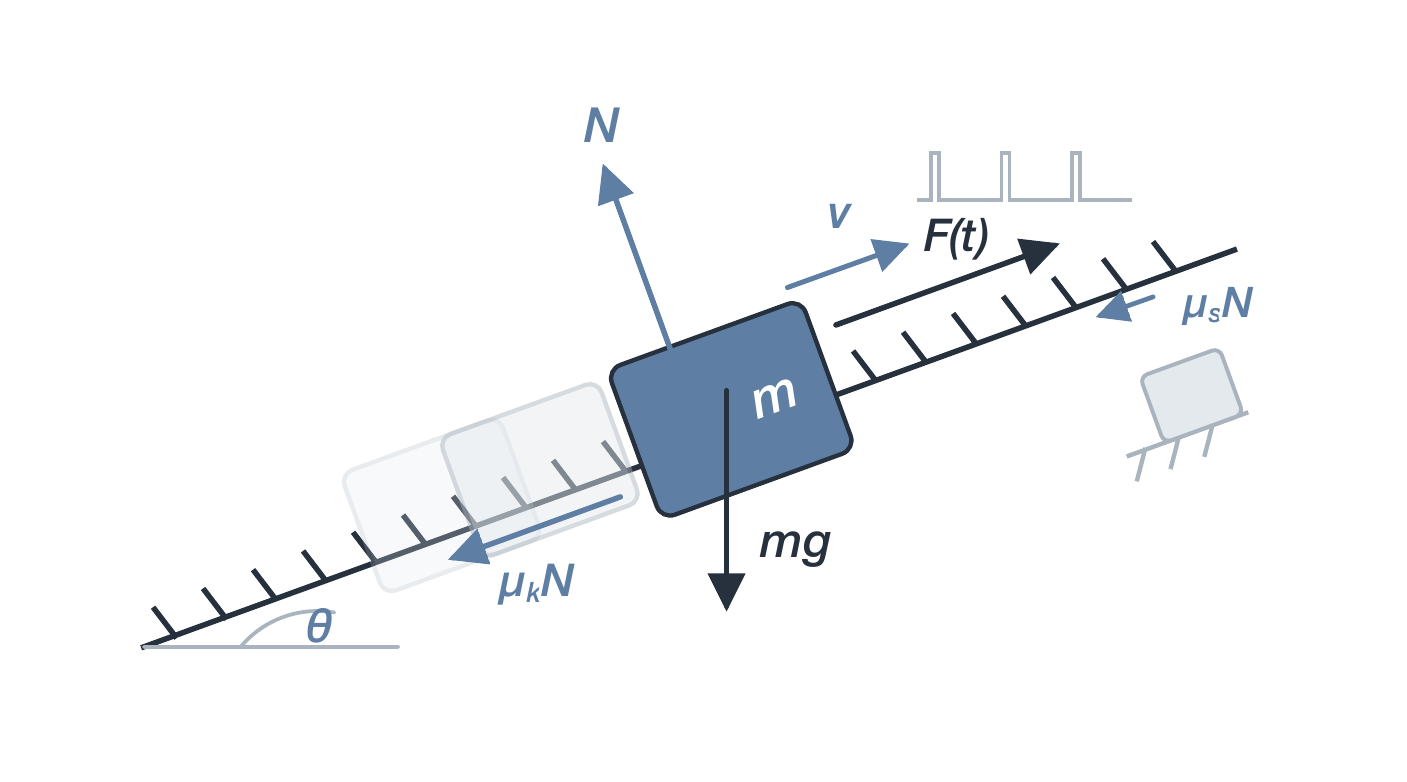}
\Description{Stylized schematic of the MassSlide simulator showing a block on an inclined plane under gravity, periodic forcing, normal force, and static and kinetic friction.}
\end{minipage}
\par

\medskip
\noindent{\large\bfseries System description\par}
\begin{DocCode}
These time series were generated by simulating a block of mass m moving along an infinitely long rigid plane inclined at an angle theta under gravitational acceleration g and Coulomb friction.
The coordinate axis is aligned with the plane.
The block is also subject to an externally applied periodic force along the plane.
The friction force acts along the plane and opposes motion.
When the block is moving, that is, when v(t) is nonzero, friction is modeled as kinetic Coulomb friction.
A breakaway static-friction threshold is also modeled: when the block is at rest, motion starts only if the net driving force along the plane exceeds a breakaway limit.
\end{DocCode}

\noindent\textbf{Observed channels:} velocity, friction force, and normal force.
\par\smallskip
\noindent\textbf{Candidate labels:} \emph{breakaway friction},
\emph{Coulomb friction}, \emph{gravity constant}, \emph{plane inclination},
and \emph{no parameter change}.

\begin{table}[!htbp]
\centering
\caption{MassSlide simulation settings used by the released trajectories.}
\label{tab:massslide-release-simulation}
\begingroup
\normalsize
\setlength{\tabcolsep}{0pt}
\setlength{\aboverulesep}{0pt}
\setlength{\belowrulesep}{2pt}
\renewcommand{\arraystretch}{1.05}
\begin{tabular*}{\columnwidth}{@{\extracolsep{\fill}}lr@{}}
\toprule
Simulation End & 10 s \\
Sampling Frequency & 50 Hz \\
Number of Rows & 499 \\
\bottomrule
\end{tabular*}
\endgroup

\end{table}
\FloatBarrier
\begin{table}[H]
\centering
\caption{Simulator-specific rejection rules for MassSlide.}
\label{tab:massslide-manual-rejection-rules}
\begingroup\small 
\setlength{\tabcolsep}{4pt} 
\setlength{\aboverulesep}{0pt} 
\setlength{\belowrulesep}{2pt} 
\renewcommand{\arraystretch}{1.08} 
\begin{tabularx}{\columnwidth}{@{}>{\raggedright\arraybackslash}p{0.26\columnwidth}>{\raggedright\arraybackslash}X@{}}
\toprule 
Applies to & Rule \\ 
\midrule 
Breakaway friction & At least one common non-time channel must have symmetric relative difference \(2|y-y_0|/(|y|+|y_0|+\epsilon_{\mathrm{SRD}})>0.2\) for five consecutive samples. \\
\addlinespace[2pt] 
Gravity & At least one of the intervention and baseline traces must contain five finite, nonzero velocity samples before and five after the reference change time, taken as the intervention time or, if unavailable, the earliest detected divergence. \\
\addlinespace[2pt] 
All remaining classes & No additional simulator-specific rule is applied. \\
\bottomrule 
\end{tabularx}
\endgroup 

\end{table}

\begin{table}[H]
\centering
\caption{Empirical MassSlide retained-pool variable summaries.}
\label{tab:massslide-release-retained-variables}
\begingroup\normalsize
\setlength{\tabcolsep}{0pt}
\renewcommand{\arraystretch}{1.05}
\begin{tabular*}{\columnwidth}{@{\extracolsep{\fill}}lrrr@{}}
\toprule
variable & min & median & max \\
\midrule
\texttt{initial\_velocity} & -5 & 1 & 12 \\
\texttt{Amplitude} & -352.94 & -27.652 & 360 \\
\texttt{Period} & 0.68753 & 1.7104 & 2.8184 \\
\texttt{mass} & 0.5 & 2.2361 & 10 \\
\texttt{gravity\_constant} & 4 & 6 & 18.432 \\
\texttt{plane\_inclination} & 5 & 20.945 & 65.938 \\
\texttt{coulomb\_friction\_coefficient} & 0.08 & 0.7746 & 6.5425 \\
\texttt{breakaway\_friction\_coefficient} & 1.095 & 6 & 42.31 \\
\bottomrule
\end{tabular*}
\par\smallskip\noindent
\begin{minipage}{0.98\columnwidth}
\footnotesize Empirical summaries are sample-weighted across every retained released sample.
\end{minipage}
\endgroup

\end{table}
\FloatBarrier
\begin{table}[H]
\centering
\caption{Exact MassSlide observation-noise coefficients.}
\label{tab:massslide-release-noise}
\begingroup\normalsize
\setlength{\tabcolsep}{0pt}
\renewcommand{\arraystretch}{1.05}
\begin{tabular*}{\columnwidth}{@{\extracolsep{\fill}}lr@{}}
\toprule
Coefficient & \(\kappa_{\mathrm{ref}}\) \\
\midrule
\texttt{friction\_base\_sigma\_scale} & 0.001 \\
\texttt{friction\_bias\_sigma\_scale} & 0.002 \\
\texttt{friction\_drift\_sigma\_scale} & 0.001 \\
\texttt{friction\_event\_sigma\_scale} & 0.02 \\
\texttt{normal\_base\_sigma\_scale} & 0.001 \\
\texttt{normal\_drift\_sigma\_scale} & 0.0002 \\
\texttt{velocity\_base\_sigma\_scale} & 0.003 \\
\texttt{velocity\_drift\_sigma\_scale} & 0.0005 \\
\texttt{velocity\_event\_sigma\_scale} & 0.01 \\
\texttt{velocity\_hetero\_sigma\_scale} & 0.003 \\
\bottomrule
\end{tabular*}
\endgroup

\end{table}

\begin{figure*}[t]
\centering
\begin{minipage}[t]{0.49\textwidth}
\centering
\includegraphics[width=\linewidth]{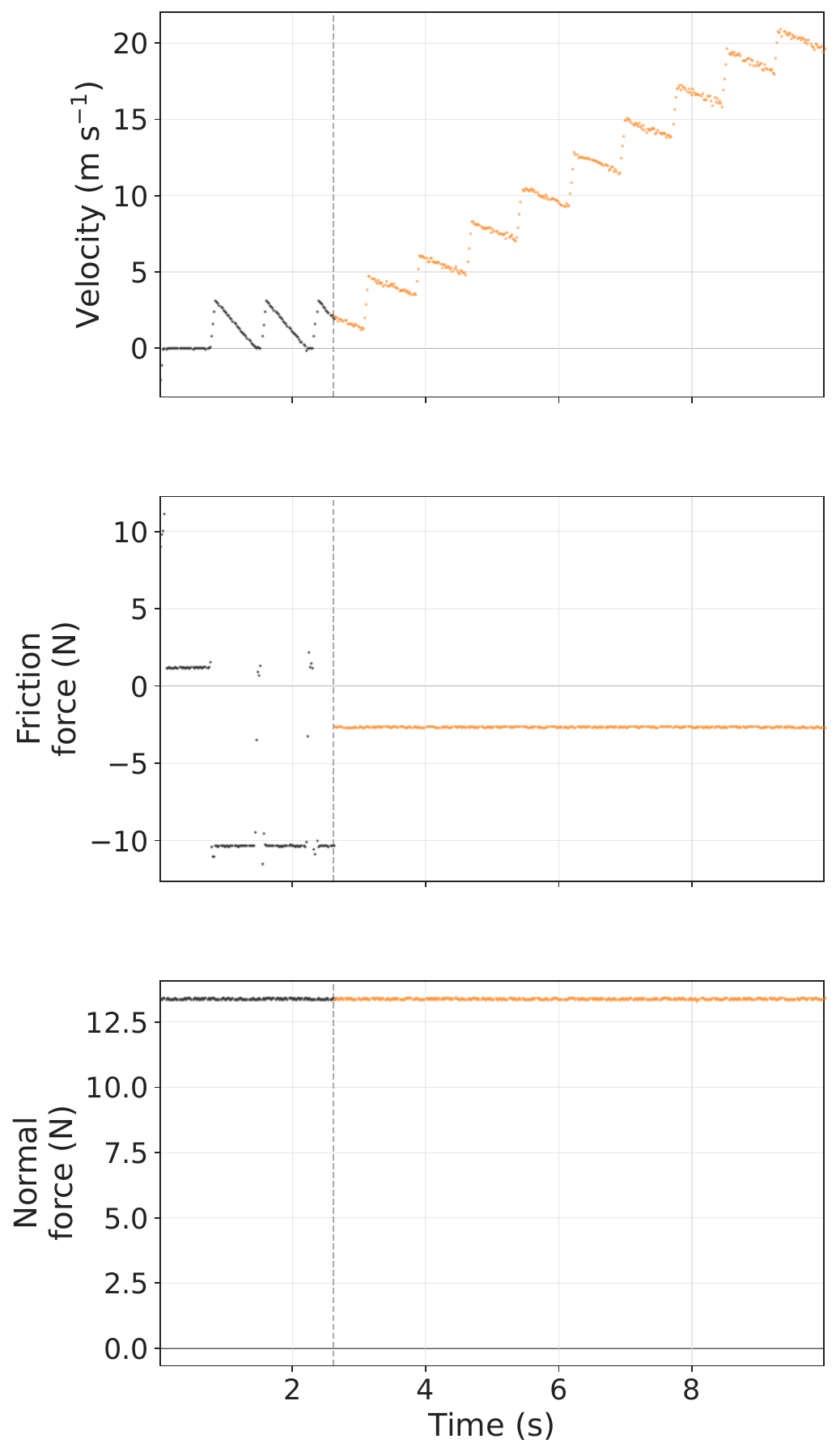}
\par\smallskip
\textbf{(a) Low observation noise}
\end{minipage}\hfill
\begin{minipage}[t]{0.49\textwidth}
\centering
\includegraphics[width=\linewidth]{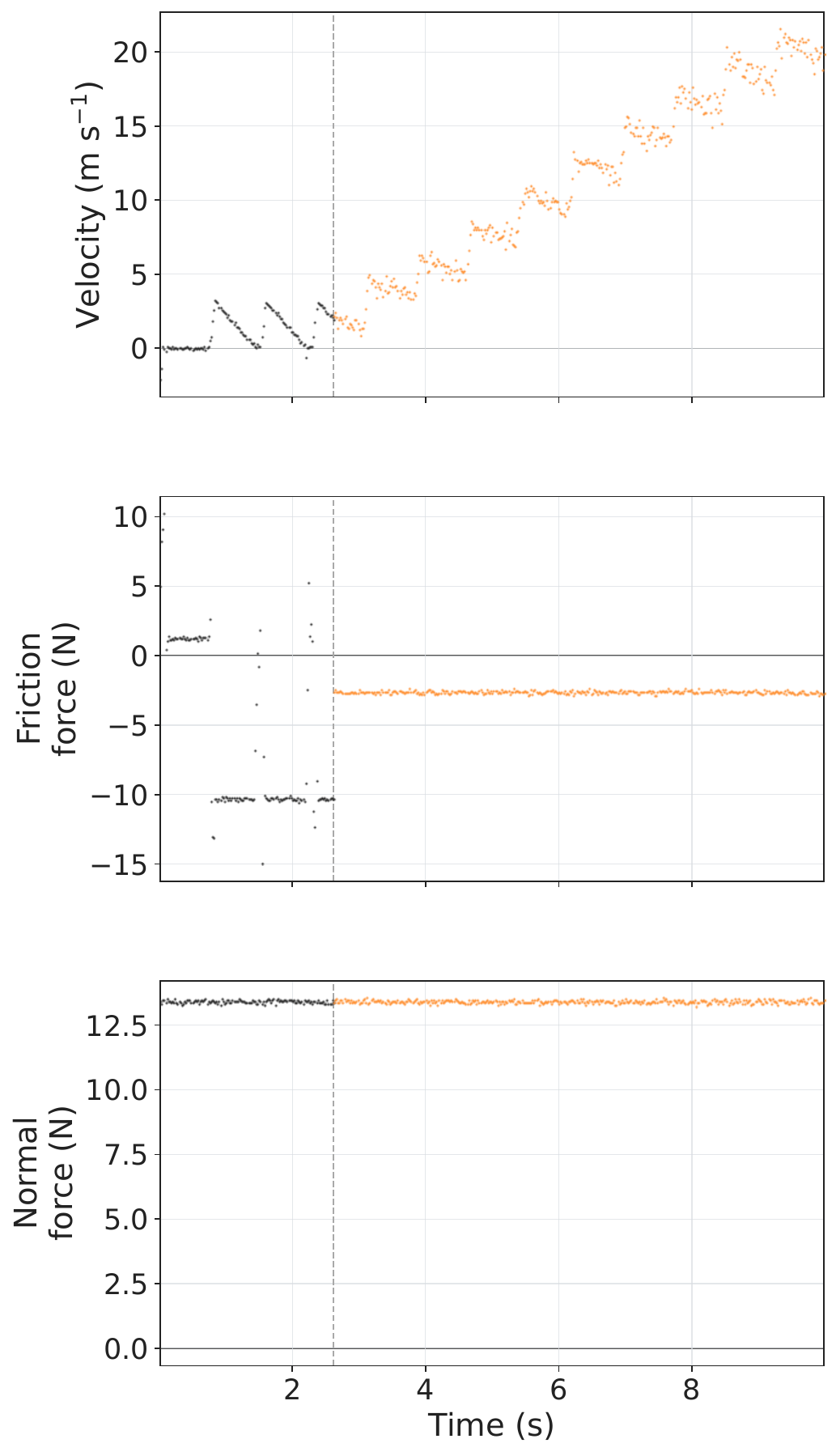}
\par\smallskip
\textbf{(b) High observation noise}
\end{minipage}
\Description{Two side-by-side sets of three aligned MassSlide time-series panels showing velocity, friction force, and normal force before and after a Coulomb-friction-coefficient intervention. The left set uses low observation noise and the right set uses high observation noise.}
\caption{Velocity, friction-force, and normal-force channels from MassSlide under (a) low and (b) high observation noise.}
\label{fig:massslide_env_noise_comparison}
\label{fig:massslide_env_low_noise}
\label{fig:massslide_env_high_noise}
\end{figure*}
\FloatBarrier

\begin{table*}[!htbp]
\centering
\caption{MassSlide released-trajectory SNR in dB as the mean and sample standard deviation.}
\label{tab:massslide-release-snr}
\begingroup
\normalsize
\providecommand{\scorepm}[2]{{\normalsize $#1 \pm #2$}}
\setlength{\tabcolsep}{0pt}
\setlength{\aboverulesep}{0pt}
\setlength{\belowrulesep}{2pt}
\renewcommand{\arraystretch}{1.05}
\begin{tabular*}{\textwidth}{@{\extracolsep{\fill}}lrr@{}}
\toprule
\texttt{Channel} & \multicolumn{1}{c}{\textbf{Low Noise}} & \multicolumn{1}{c}{\textbf{High Noise}} \\
\midrule
\texttt{friction\_force} & \scorepm{42.03}{9.75} & \scorepm{29.99}{9.75} \\
\texttt{mass\_velocity} & \scorepm{35.64}{6.58} & \scorepm{23.60}{6.58} \\
\texttt{normal\_force} & \scorepm{59.90}{0.40} & \scorepm{47.86}{0.40} \\
\bottomrule
\end{tabular*}

\endgroup

\end{table*}
\FloatBarrier
\flushbottom

\clearpage
\section{Additional Details on Solvability Filtering}
\label{app:solvability-filtering}
We apply the solvability filtering procedure to noise-free intervention trajectories before sampling observation noise.
Let \(\mathcal{T}\) denote the observed time indices, and let \(q\in\{\mathrm{NI},\mathrm{SC}\}\) identify the no-intervention and static-change probes.
For non-impulse channels, we define the comparison window as
\[
\mathcal{I}_q
=
\begin{cases}
\{t\in\mathcal{T}:t\geq t_{\mathrm{int}}\},
& q=\mathrm{NI},\\
\mathcal{T},
& q=\mathrm{SC},
\end{cases}
\qquad
T_q=|\mathcal{I}_q|.
\]
The no-intervention probe is compared only after the intervention, whereas the static-change probe is compared over the full trajectory.
For each channel \(c\), let \(\mathcal{I}_{c,q}\) denote its comparison window. 
For non-impulse channels, we set \(\mathcal{I}_{c,q}=\mathcal{I}_q\).

For impulse channels, such as the contact-force channels in \texttt{BallDrop} and \texttt{BounceBall}, we instead use the portions of \(\mathcal{I}_q\) lying within \(30\) ms of impacts detected in either the target or probe trajectory. 
Let
\[
T_{c,q}=|\mathcal{I}_{c,q}|.
\]

For channel \(c\) and probe \(q\), the trajectory difference is
\[
\Delta_{c,q}
=
\sqrt{
\frac{1}{T_{c,q}}
\sum_{t\in\mathcal{I}_{c,q}}
\left(
x^{\mathrm{target}}_{c,t}
-
x^{\mathrm{probe},q}_{c,t}
\right)^2
}.
\]
We normalize this difference by the expected corruption produced by an observation model with noise level \(\kappa\), evaluated on the same clean target trajectory and comparison window:
\[
\sigma_{c,q}(\kappa)
=
\sqrt{
\mathbb{E}_{\tilde{\mathbf{x}}\sim
p_{\kappa}(\cdot\mid\mathbf{x}^{\mathrm{target}})}
\left[
\frac{1}{T_{c,q}}
\sum_{t\in\mathcal{I}_{c,q}}
\left(
\tilde{x}_{c,t}
-
x^{\mathrm{target}}_{c,t}
\right)^2
\right]
}.
\]
The resulting dimensionless effect-to-noise ratio and its aggregation across the observed channels are
\[
R_{c,q}(\kappa)
=
\frac{\Delta_{c,q}}{\sigma_{c,q}(\kappa)+\varepsilon},
\qquad
R_q(\kappa)=\max_c R_{c,q}(\kappa),
\]
where \(\varepsilon>0\) provides numerical stability.
We estimate the expectation with deterministic noise seeds \(0,\ldots,N-1\).
Candidates are first screened using \(N=128\) high-noise realizations.
Every screen-passing candidate is then recomputed using \(N=1024\) realizations, and it is retained only if it passes at both stages.
For the filtering procedure, we evaluate this criterion at \(\kappa=\kappa_{\mathrm{high}}\) and use the single global threshold \(\rho=2\).
A target and probe \(q\) are considered distinguishable when
\[
R_q(\kappa_{\mathrm{high}})>\rho,
\]
meaning that, in at least one observed channel, the effect-to-noise ratio exceeds two.

For each intervention candidate, we evaluate the two probe comparisons independently and retain the candidate only when
\[
R_{\mathrm{NI}}(\kappa_{\mathrm{high}})>\rho
\qquad\text{and}\qquad
R_{\mathrm{SC}}(\kappa_{\mathrm{high}})>\rho.
\]

The simulator-specific rejection rules described in \Cref{app:simulator-catalog} are applied in addition to this criterion.

\clearpage
\section{Additional Details about Experimental Setup}
\label{app:experimental-setup-details}

\subsection{Data Generation and Sample Selection}
Initial configurations, each comprising an initial state and a complete set of baseline parameter values, were constructed from the simulator-specific operating ranges reported in \Cref{app:simulator-catalog}.
For each initial configuration, we generated multiple candidates by varying the intervention class, intervention time, and post-intervention parameter value.
Each candidate was subsequently processed by the documented rejection pipeline.

To reduce the risk that the pre-intervention configuration alone reveals the ground-truth class, we retained an initial configuration only if at least two distinct root-cause interventions passed the filtering procedure.
No-intervention samples were also generated from these retained initial configurations.

Additionally, test samples within the same batch and, where applicable, labeled support samples were selected from distinct initial configurations.
Thus, no two samples in the same episode shared the same initial state and baseline parameter values, reducing the risk of biasing the agent.

The held-out test set comprises 142 samples per class.
All datasets and sample assignments were frozen before any agent evaluation was conducted.

\subsection{Released-test-set accounting and construction}
\label{app:environment-release-statistics}
All statistics in this catalog are recomputed from the released \texttt{questions.json}, \texttt{sample\_manifest.json}, \texttt{model\_record.json}, clean Parquet trajectories, and released noise implementations.
Empirical parameter and intervention summaries use the full retained environment pool rather than an illustrative subset.

\subsection{Available Platforms}
\label{app:agent-platforms-and-profiles}

\begingroup
\normalsize
\setlength{\tabcolsep}{4pt}
\renewcommand{\arraystretch}{1.1}
\begin{table}[!htbp]
\centering
\caption{Available platforms and their installation requirements.}
\begin{tabularx}{\columnwidth}{@{}P{0.29\columnwidth}>{\raggedright\arraybackslash\normalsize}X@{}}
\toprule
\textbf{Platform} & \textbf{Pinned npm package} \\
\midrule
\code{codex} & \code{@openai/codex@0.124} \\
\code{gemini-cli} & \code{@google/gemini-cli@0.35.3} \\
\code{claude-code} & \code{@anthropic-ai/claude-code@2.1.96} \\
\code{opencode} & \code{opencode-ai@1.14.29} \\
\bottomrule
\end{tabularx}
\end{table}
\endgroup

\subsection{Available Agents}
Only the profiles used in the paper are shown.

\begingroup
\normalsize
\setlength{\tabcolsep}{3pt}
\renewcommand{\arraystretch}{1.08}
\begin{table}[!htbp]
\centering
\caption{Agent profile catalog.}
\begin{tabularx}{\columnwidth}{@{}>{\raggedright\arraybackslash}X P{0.25\columnwidth}cc@{}}
\toprule
\textbf{Agent} & \textbf{Platform} & \textbf{Reasoning Effort} & \textbf{Img.} \\
\midrule
\code{claude-opus-4.6} & \code{claude-code} & high & Y \\
\code{gemini-3.1-pro} & \code{gemini-cli} & high & Y \\
\code{gpt-5.5} & \code{codex} & high & Y \\
\code{minimax-m2.7} & \code{opencode} & -- & N \\
\bottomrule
\end{tabularx}
\end{table}
\endgroup

\FloatBarrier

\ifdefined\tracebenchDocsCombinedAppendixMode
\def\tracebenchAppendixRuntimeLibrariesStart{%
  \docTitle{Appendix: Agentic Runtime Libraries}%
  \phantomsection\label{docstart:appendix_runtime_libraries}%
  \section*{Agentic Runtime Libraries}%
}
\def\tracebenchAppendixRuntimeLibrariesEnd{}
\else\ifdefined\tracebenchCombinedAppendixMode
\def\tracebenchAppendixRuntimeLibrariesStart{}
\def\tracebenchAppendixRuntimeLibrariesEnd{}
\else
\documentclass[11pt]{article}
\usepackage{styles/manual_docs_style}
\usepackage{booktabs}

\def\tracebenchAppendixRuntimeLibrariesStart{%
  \begin{document}%
  \docTitle{Appendix: Agentic Runtime Libraries}%
  \phantomsection\label{docstart:appendix_runtime_libraries}%
  \section*{Agentic Runtime Libraries}%
}
\def\tracebenchAppendixRuntimeLibrariesEnd{\end{document}}
\fi\fi
\def\tracebenchRuntimeLibrariesIntro{%
Agent episodes run in an isolated container based on
\code{ghcr.io/laude-institute/t-bench/python-3-13:20250620}.
Internet access is disabled except for the model API proxy.
The runtime also includes \code{uv} version \code{0.8.15} and the system
utilities \code{tmux}, \code{asciinema}, and \code{procps}.
The Python packages explicitly installed in the agent runtime are listed below.
}
\tracebenchAppendixRuntimeLibrariesStart

\ifdefined\tracebenchCombinedAppendixMode
\subsection{Agentic Runtime Libraries}
\label{app:agent-runtime-libraries}
\tracebenchRuntimeLibrariesIntro
\medskip
\else
\tracebenchRuntimeLibrariesIntro
\fi

\begingroup
\normalsize
\setlength{\tabcolsep}{6pt}
\renewcommand{\arraystretch}{1.08}
\begin{table}[!htbp]
\centering
\caption{Python packages installed in the agent runtime.}
\begin{tabularx}{\columnwidth}{@{}>{\raggedright\arraybackslash}X>{\normalsize}P{0.32\columnwidth}@{}}
\toprule
\textbf{Package} & \textbf{Version} \\
\midrule
\code{contourpy} & \code{1.3.3} \\
\code{cycler} & \code{0.12.1} \\
\code{fonttools} & \code{4.61.1} \\
\code{iniconfig} & \code{2.3.0} \\
\code{joblib} & \code{1.5.3} \\
\code{kiwisolver} & \code{1.4.9} \\
\code{llvmlite} & \code{0.46.0} \\
\code{matplotlib} & \code{3.10.8} \\
\code{numba} & \code{0.63.1} \\
\code{numpy} & \code{2.3.5} \\
\code{packaging} & \code{26.0} \\
\code{pandas} & \code{3.0.0} \\
\code{pillow} & \code{12.1.0} \\
\code{pluggy} & \code{1.6.0} \\
\code{pyarrow} & \code{23.0.0} \\
\code{Pygments} & \code{2.19.2} \\
\code{pyparsing} & \code{3.3.2} \\
\code{pytest} & \code{8.4.1} \\
\code{python-dateutil} & \code{2.9.0.post0} \\
\code{ruptures} & \code{1.1.10} \\
\code{scikit-learn} & \code{1.8.0} \\
\code{scipy} & \code{1.17.0} \\
\code{six} & \code{1.17.0} \\
\code{threadpoolctl} & \code{3.6.0} \\
\code{tslearn} & \code{0.7.0} \\
\bottomrule
\end{tabularx}
\end{table}
\endgroup

\tracebenchAppendixRuntimeLibrariesEnd

\FloatBarrier

\clearpage
\begingroup
\raggedbottom
\section{Evaluation Costs}
\label[appendix]{app:evaluation-costs}

\subsection{Agent Pricing}
\label[appendix]{app:agent-pricing}

\begin{table}[H]
  \centering
  \normalsize
  \setlength{\tabcolsep}{3.5pt}
  \renewcommand{\arraystretch}{1.08}
  \caption{Agent prices used to compute evaluation costs, in USD per 1M tokens.}
  \label{tab:app-agent-pricing}
  \begin{tabularx}{\linewidth}{@{}>{\raggedright\arraybackslash}X*{3}{>{\normalsize}r}@{}}
    \toprule
    \textbf{Agent} & \textbf{Input} & \textbf{Cached} & \textbf{Output} \\
    \midrule
    \code{claude-opus-4.6} & 5 & 0.5 & 25 \\
    \code{gemini-3.1-pro} & 2 & 0.2 & 12 \\

    \code{gpt-5.5} & 5 & 0.5 & 30 \\
    \code{minimax-m2.7} & 0.3 & 0.059 & 1.20 \\
    \bottomrule
  \end{tabularx}
\end{table}

\FloatBarrier

\subsection{Accuracy versus Evaluation Cost}
\label[appendix]{app:accuracy-vs-cost}

\Cref{fig:accuracy-vs-cost-balldrop,fig:accuracy-vs-cost-bounceball,fig:accuracy-vs-cost-massslide} show episode accuracy versus evaluation cost separately for each \termSimulator and each main-experiment condition.
Each point shows the mean episode accuracy and mean evaluation cost for one model across the five repetition seeds; horizontal and vertical error bars show the corresponding sample standard deviations.

\begin{figure}[H]
  \centering
  \includegraphics[width=0.95\linewidth]{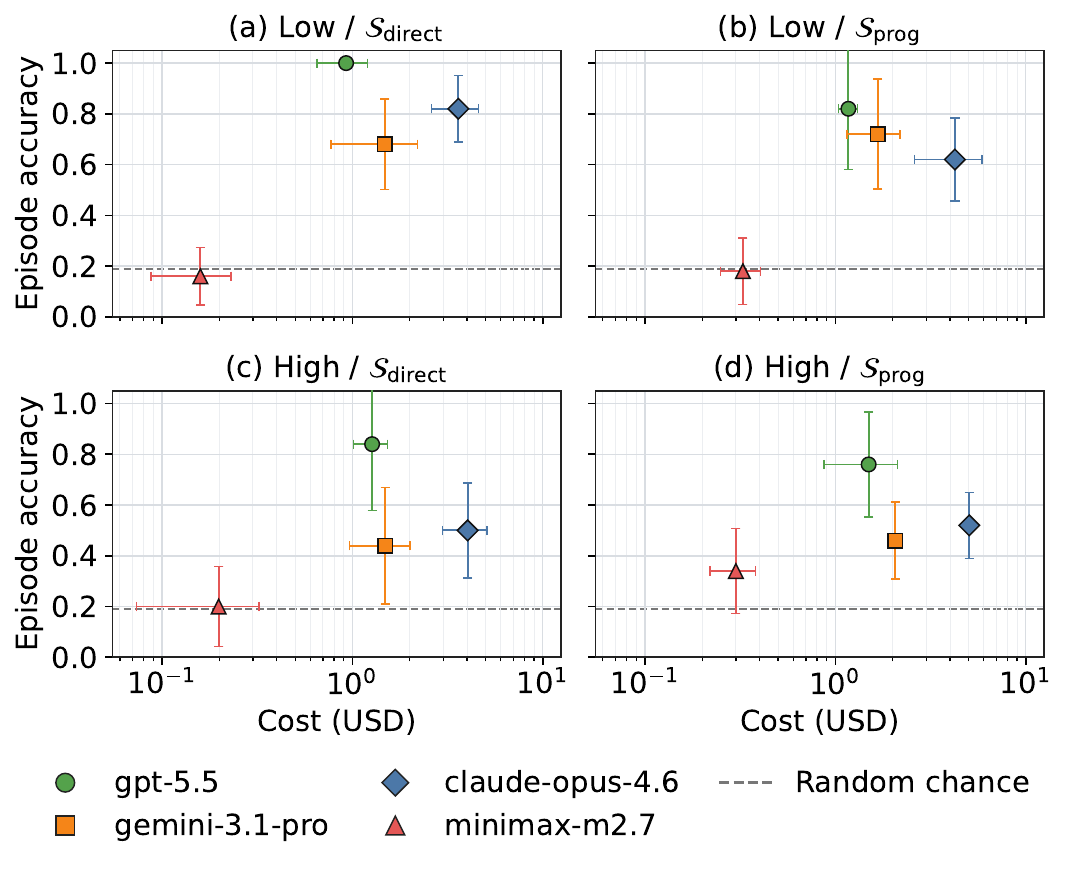}
  \Description{Scatter plot of episode top-1 accuracy versus evaluation cost for BallDrop across models and experimental conditions.}
  \caption{Episode top-1 accuracy versus mean evaluation cost for \texttt{BallDrop} under the four main-experiment conditions.}
  \label{fig:accuracy-vs-cost-balldrop}
\end{figure}

\begin{figure}[H]
  \centering
  \includegraphics[width=0.95\linewidth]{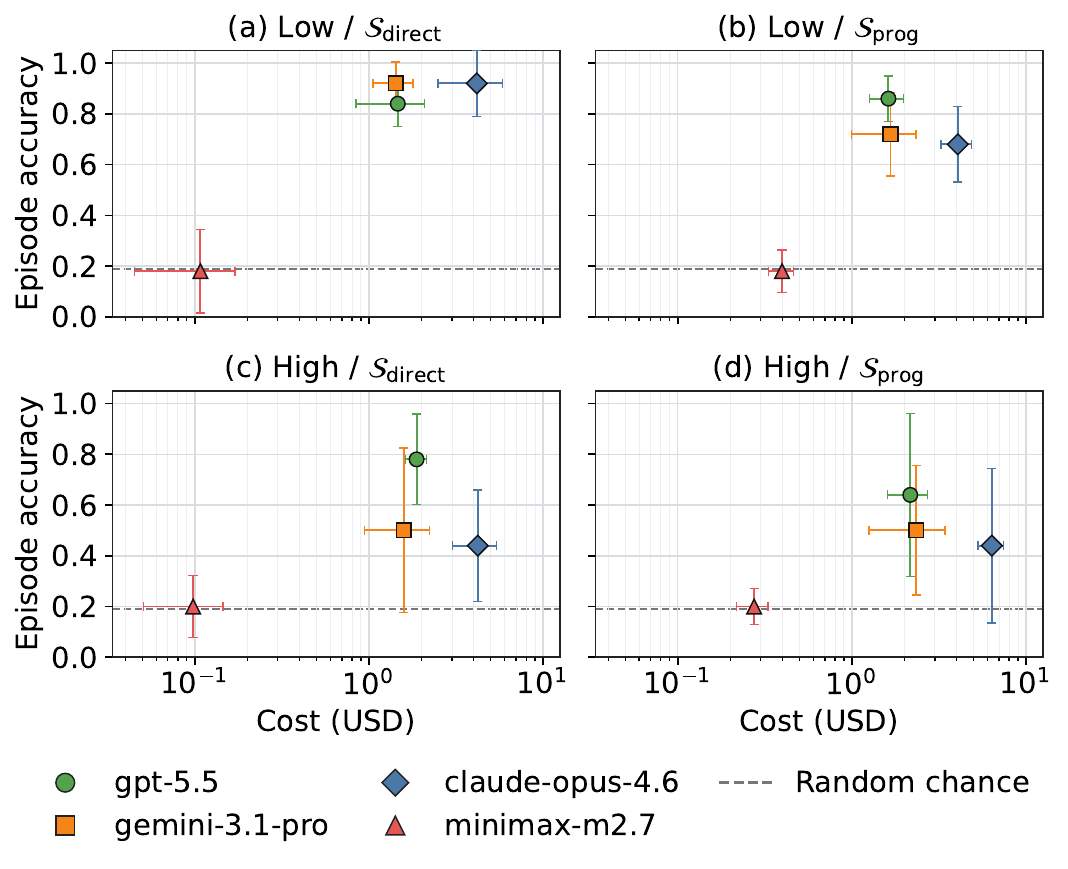}
  \Description{Scatter plot of episode top-1 accuracy versus evaluation cost for BounceBall across models and experimental conditions.}
  \caption{Episode top-1 accuracy versus mean evaluation cost for \texttt{BounceBall} under the four main-experiment conditions.}
  \label{fig:accuracy-vs-cost-bounceball}
\end{figure}

\begin{figure}[H]
  \centering
  \includegraphics[width=0.95\linewidth]{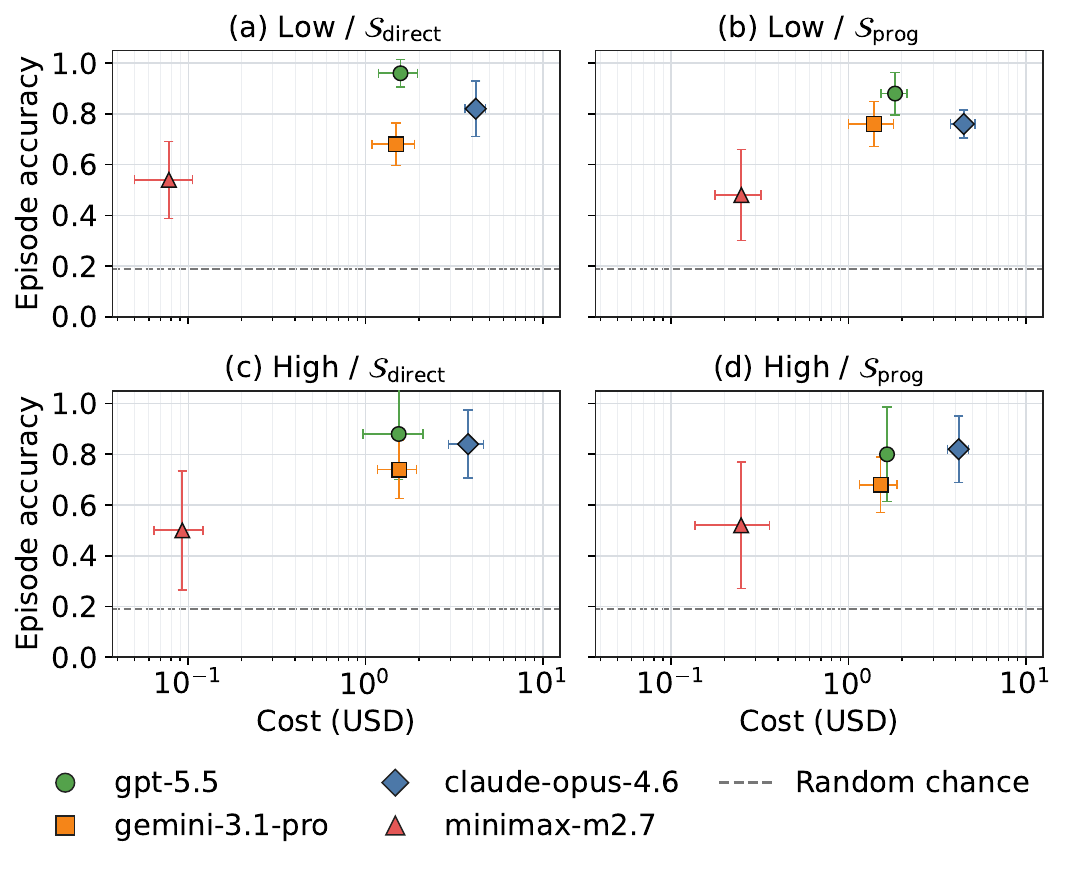}
  \Description{Scatter plot of episode top-1 accuracy versus evaluation cost for MassSlide across models and experimental conditions.}
  \caption{Episode top-1 accuracy versus mean evaluation cost for \texttt{MassSlide} under the four main-experiment conditions.}
  \label{fig:accuracy-vs-cost-massslide}
\end{figure}


\clearpage
\endgroup

\twocolumn[
\begin{minipage}{\textwidth}
\section{Classical Time-Series Baselines}
\label{app:classical-baselines}
\centering
\captionof{table}{Visible-test accuracy of six learned classical time-series baselines, including privileged variants that subtract the clean paired no-intervention trajectory; uniform-random-guessing accuracy for \texttt{BallDrop}/\texttt{BounceBall}/\texttt{MassSlide} is 0.200/0.167/0.200 (aggregate: 0.189).}
\label{tab:classical-baselines}
\begingroup
\normalsize
\providecommand{\scorepm}[2]{{\normalsize $#1 \pm #2$}}
\setlength{\tabcolsep}{4pt}
\setlength{\aboverulesep}{0pt}
\setlength{\belowrulesep}{2pt}
\renewcommand{\arraystretch}{1.0}

\begin{tabularx}{\textwidth}{@{}>{\raggedright\arraybackslash}X*{4}{>{\normalsize}c}@{}}
\toprule
Method & \texttt{BallDrop} & \texttt{BounceBall} & \texttt{MassSlide} & Aggregate \\

\addlinespace[2pt]\midrule
\multicolumn{5}{c}{\textbf{Low noise ($\kappa_{\mathrm{low}}$)}} \\
\midrule

\texttt{Euclidean 1-NN} & \scorepm{0.140}{0.089} & \scorepm{0.340}{0.152} & \scorepm{0.140}{0.055} & \scorepm{0.207}{0.099} \\
\texttt{Euclidean 1-NN (privileged)} & \scorepm{0.420}{0.148} & \scorepm{0.360}{0.114} & \scorepm{0.480}{0.192} & \scorepm{0.420}{0.143} \\
\texttt{DTW 1-NN} & \scorepm{0.220}{0.148} & \scorepm{0.240}{0.134} & \scorepm{0.220}{0.110} & \scorepm{0.227}{0.122} \\

\texttt{DTW 1-NN (privileged)} & \scorepm{0.520}{0.130} & \scorepm{0.360}{0.230} & \scorepm{0.580}{0.192} & \scorepm{0.487}{0.175} \\
\texttt{MiniRocket+ridge} & \scorepm{0.260}{0.114} & \scorepm{0.320}{0.110} & \scorepm{0.320}{0.130} & \scorepm{0.300}{0.110} \\
\texttt{MiniRocket+ridge (privileged)} & \scorepm{0.600}{0.158} & \scorepm{0.720}{0.192} & \scorepm{0.500}{0.122} & \scorepm{0.607}{0.148} \\
\addlinespace[2pt]\midrule
\multicolumn{5}{c}{\textbf{High noise ($\kappa_{\mathrm{high}}$)}} \\
\midrule

\texttt{Euclidean 1-NN} & \scorepm{0.140}{0.089} & \scorepm{0.340}{0.152} & \scorepm{0.140}{0.055} & \scorepm{0.207}{0.099} \\
\texttt{Euclidean 1-NN (privileged)} & \scorepm{0.420}{0.148} & \scorepm{0.360}{0.152} & \scorepm{0.460}{0.167} & \scorepm{0.413}{0.144} \\
\texttt{DTW 1-NN} & \scorepm{0.220}{0.148} & \scorepm{0.260}{0.114} & \scorepm{0.220}{0.110} & \scorepm{0.233}{0.116} \\

\texttt{DTW 1-NN (privileged)} & \scorepm{0.500}{0.141} & \scorepm{0.380}{0.179} & \scorepm{0.440}{0.089} & \scorepm{0.440}{0.131} \\
\texttt{MiniRocket+ridge} & \scorepm{0.280}{0.110} & \scorepm{0.280}{0.130} & \scorepm{0.280}{0.148} & \scorepm{0.280}{0.121} \\
\texttt{MiniRocket+ridge (privileged)} & \scorepm{0.600}{0.212} & \scorepm{0.680}{0.192} & \scorepm{0.400}{0.071} & \scorepm{0.560}{0.158} \\
\bottomrule
\end{tabularx}
\endgroup


\end{minipage}
\vspace{0.5\baselineskip}
]

We compare the agent results with six learned non-agentic baselines and analytic uniform random guessing on the matched visible-test direct-answer episodes under low and high observation noise.
For each \termSimulator and seed, every learned baseline fits only the three labeled examples per class shown to the agents and predicts the same ten test trajectories.
The six learned methods pair ordinary and privileged variants of Euclidean 1-nearest-neighbor (1-NN), multivariate dynamic time warping (DTW) 1-NN, and MiniRocket with a ridge classifier~\cite{dempster2021minirocket}.
Each privileged input is the observed intervention trajectory minus its clean paired no-intervention trajectory, channel by channel.
We retain each original time grid, exclude its time column, and perform no interpolation or standardization.
After noise addition and privileged subtraction, we divide only \code{Hard_Stop_f} for \texttt{BallDrop} and \code{Right_Hard_Stop_f} for \texttt{BounceBall} by that trajectory's maximum absolute impulse force when nonzero.
Euclidean 1-NN uses root-mean-square distance on the flattened multichannel trajectory.
DTW 1-NN uses dependent multivariate DTW with squared-Euclidean local cost and an unconstrained warping path.
MiniRocket uses 10,000 kernels, a fixed random seed, and aeon's \(10^{-7}\) constant-channel threshold, while ridge cross-validation is confined to the labeled support set.
Ties for the nearest-neighbor methods are resolved deterministically by distance, class index, and support-sample identifier.

\clearpage
\begin{figure*}[p]
  \section{Direct-Answer Confusion Matrices}
  \label{app:direct-confusion-matrices}
  \centering
  \includegraphics[
    width=\textwidth,
    height=0.87\textheight,
    keepaspectratio
  ]{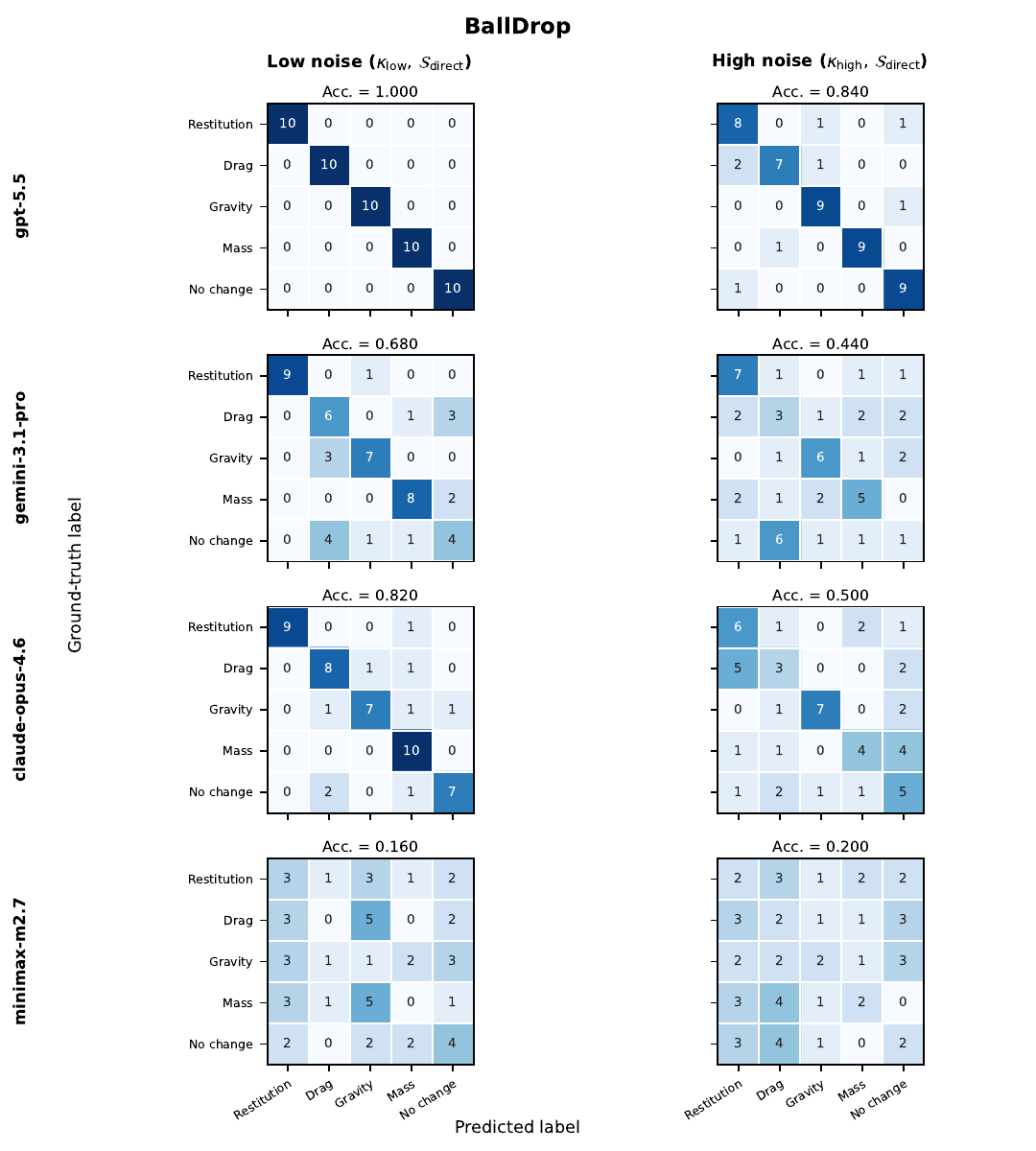}
  \Description{Eight BallDrop confusion matrices arranged in four agent rows
  and two noise-level columns. Rows within each matrix are ground-truth labels,
  columns are predicted labels, cells contain raw counts, and color encodes
  the row-normalized fraction.}
  \caption{BallDrop confusion matrices for
  \(\mathcal{S}_{\mathrm{direct}}\), reported separately for every agent under
  \(\kappa_{\mathrm{low}}\) and \(\kappa_{\mathrm{high}}\).}
  \label{fig:direct-confusion-balldrop}
\end{figure*}

\begin{figure*}[p]
  \centering
  \includegraphics[
    width=\textwidth,
    height=0.87\textheight,
    keepaspectratio
  ]{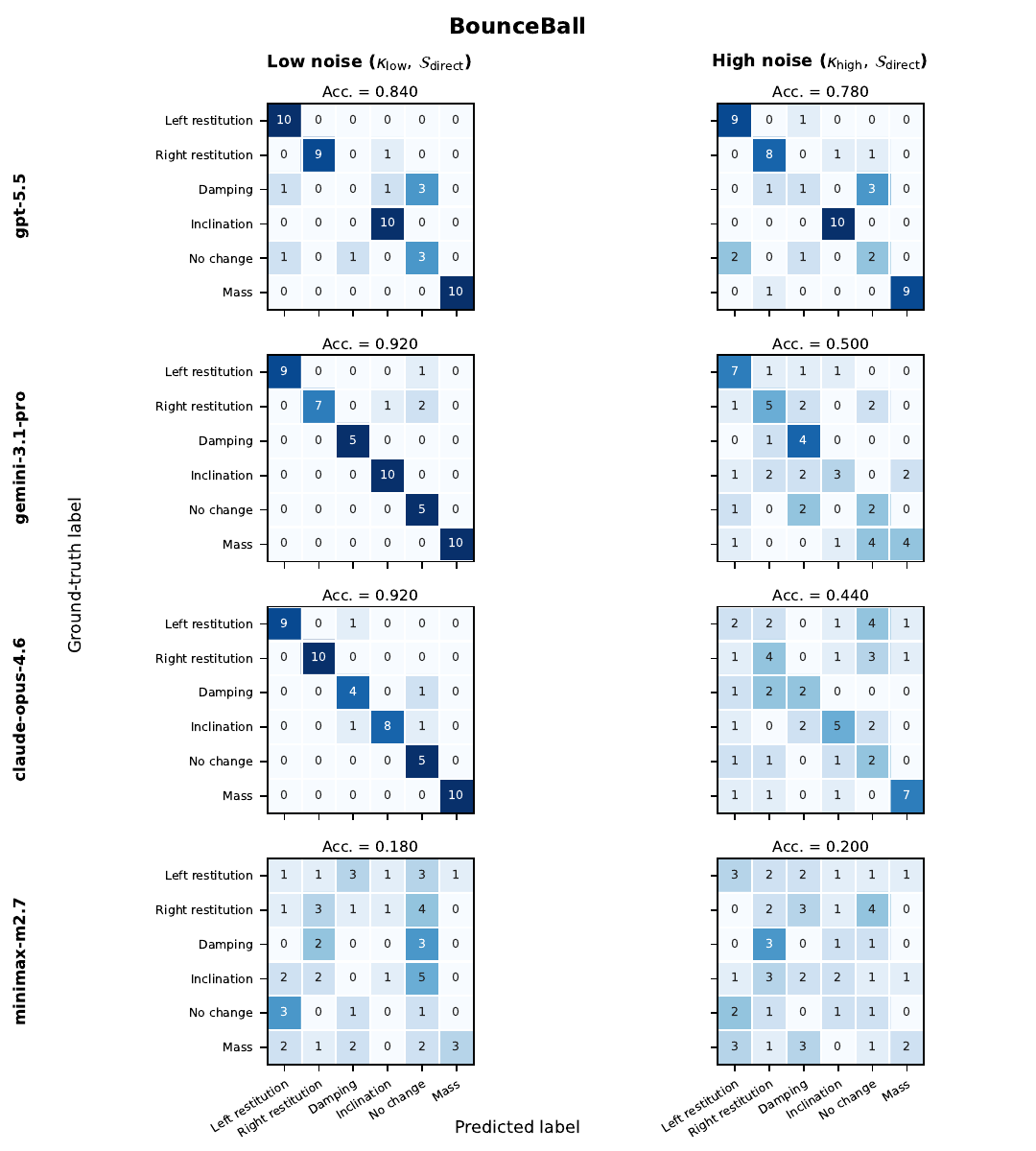}
  \Description{Eight BounceBall confusion matrices arranged in four agent rows
  and two noise-level columns. Rows within each matrix are ground-truth labels,
  columns are predicted labels, cells contain raw counts, and color encodes
  the row-normalized fraction.}
  \caption{BounceBall confusion matrices for
  \(\mathcal{S}_{\mathrm{direct}}\), reported separately for every agent under
  \(\kappa_{\mathrm{low}}\) and \(\kappa_{\mathrm{high}}\).}
  \label{fig:direct-confusion-bounceball}
\end{figure*}

\begin{figure*}[p]
  \centering
  \includegraphics[
    width=\textwidth,
    height=0.87\textheight,
    keepaspectratio
  ]{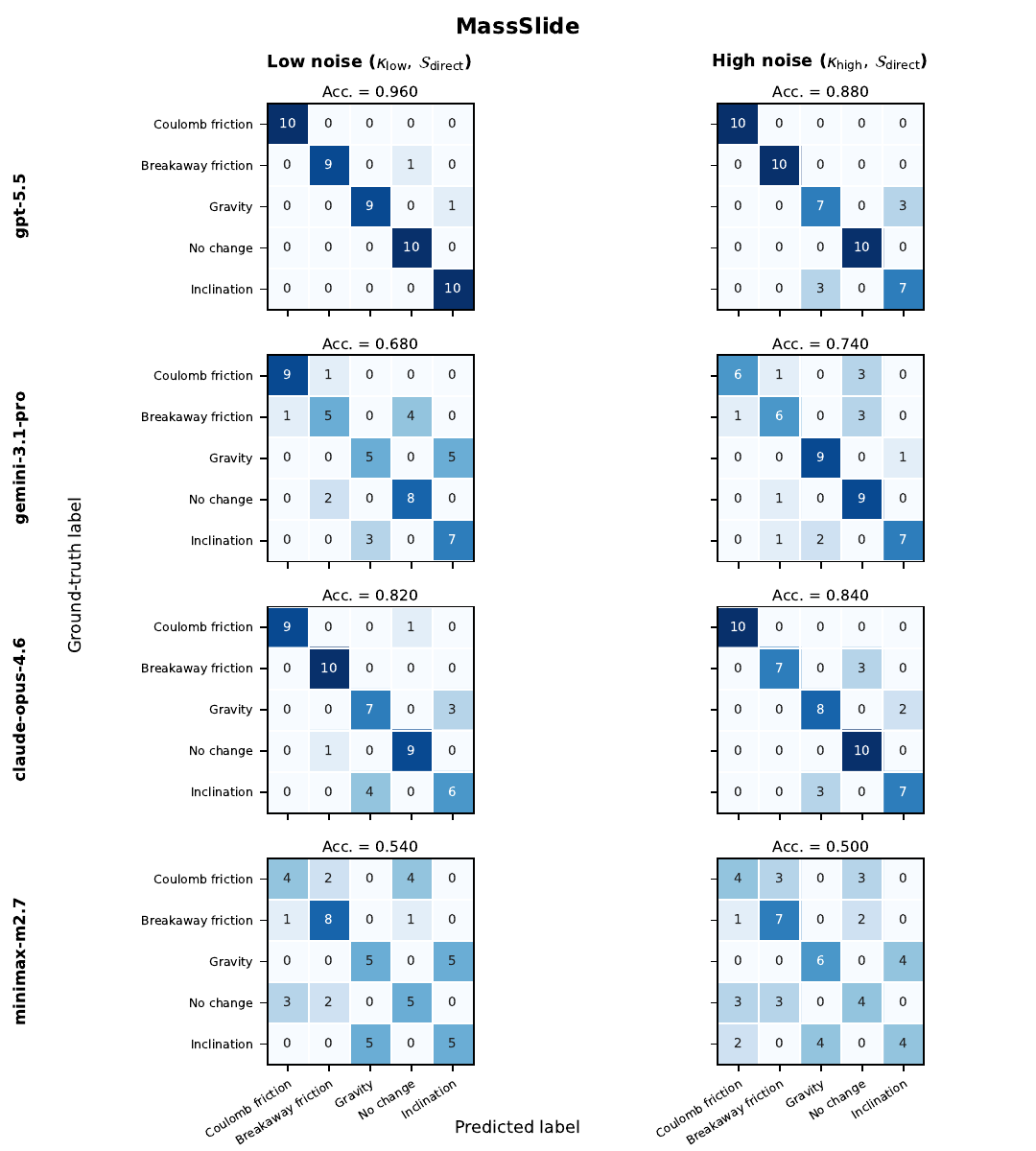}
  \Description{Eight MassSlide confusion matrices arranged in four agent rows
  and two noise-level columns. Rows within each matrix are ground-truth labels,
  columns are predicted labels, cells contain raw counts, and color encodes
  the row-normalized fraction.}
  \caption{MassSlide confusion matrices for
  \(\mathcal{S}_{\mathrm{direct}}\), reported separately for every agent under
  \(\kappa_{\mathrm{low}}\) and \(\kappa_{\mathrm{high}}\).}
  \label{fig:direct-confusion-massslide}
\end{figure*}

\FloatBarrier

\clearpage
\section{Metric definitions and computation}
\label{app:metric-computation}
This section defines the trajectory-derived metrics reported in the main text.

\subsection{Interaction and Python-use metrics}

\paragraph{Agent iterations.}
The number of LLM calls in a single trajectory.
An iteration can contain text, reasoning, and zero or more tool calls.

\paragraph{Tool calls.}
The total number of well-formed tool calls in an episode.

\paragraph{Python calls.}
The total number of parsed Python interpreter invocations in tool commands.
Inline \texttt{python -c} programs, Python heredocs, module invocations, and executions of \texttt{.py} files are counted; repeated executions of the same source are separate calls.
Displaying Python source without executing it does not add a call.

\paragraph{Python statements.}
The total number of Python abstract-syntax-tree statement nodes across the distinct Python source versions first displayed in the trajectory.

\subsection{Plot and numerical-inspection metrics}

\paragraph{Plots generated.}
The number of image-like artifacts created in the episode artifact directory, including common raster, vector, and PDF formats.
Each distinct generated file contributes one, irrespective of how often it is later opened.

\paragraph{Plots inspected.}
The number of images generated by the LLM agent that are subsequently read, opened, or viewed successfully with an image-capable tool.

\paragraph{Numeric values printed.}
This metric counts the numeric literals exposed to the agent through tool output during an episode.
It includes integers, decimals, scientific notation, numeric entries in arrays and tables, and non-finite forms such as \texttt{nan} and \texttt{inf}.
It excludes numbers that occur in source code, tool-wrapper metadata, process identifiers, warnings and tracebacks, synthetic file-reader line numbers, sample and filesystem identifiers, and embedded image payloads.
Consequently, this metric measures numerical evidence exposed through tool output rather than the number of values present in the input time-series files.

\paragraph{Console output token share.}
This metric estimates the percentage of cumulative input-context tokens attributable to console output.

\subsection{Token, cost, and context metrics}

\paragraph{Total tokens.}
The total number of prompt tokens processed plus completion tokens generated across an episode.

\paragraph{Cost.}
Evaluation cost is reported in U.S. dollars and is computed from the recorded prompt, cached-prompt, and completion-token totals using the per-million-token prices configured for the corresponding agent profile:
\begin{equation}
    C
    = \frac{(T_{\mathrm{prompt}}-T_{\mathrm{cached}})p_{\mathrm{in}}
      + T_{\mathrm{cached}}p_{\mathrm{cached}}
      + T_{\mathrm{completion}}p_{\mathrm{out}}}{10^{6}}.
\end{equation}
The per-agent input, cached-input, and output prices used in this calculation are reported in \Cref{tab:app-agent-pricing}.

\paragraph{Total time.}
Total time measures active agent-system wall-clock time, including model inference and tool execution.

\clearpage
\begingroup
\makeatletter
\renewenvironment{table*}[1][p]{%
  \@dblfloat{table}[#1]%
  \section{Paired Statistical Analysis and Uncertainty}
  \label{app:statistical-analysis}%
}{%
  \end@dblfloat
}
\makeatother
\begin{table*}[p]
\centering
\caption{Complete paired episode-level differences, reported in percentage points with confidence intervals obtained by exhaustive sign-flip inversion.
For agent comparisons, differences are calculated within each simulator--seed pair and then averaged across the low- and high-noise conditions.
For condition comparisons, differences are averaged across the three closed-weight agents; \texttt{minimax-m2.7} is excluded.
W/T/L (win/tie/loss) is determined from these paired differences.}
\label{tab:paired-contrasts}
\begingroup
\normalsize
\setlength{\tabcolsep}{2pt}
\renewcommand{\arraystretch}{0.80}
\begin{tabularx}{\textwidth}{@{}>{\raggedright\arraybackslash}X>{\centering\arraybackslash}p{0.29\textwidth}>{\centering\arraybackslash}p{0.09\textwidth}@{}}
\toprule
Contrast & Effect [95\% CI] & W/T/L \\
\midrule
\multicolumn{3}{l}{\textbf{Agent comparisons (low/high noise averaged)}}\\
gpt-5.5 - gemini-3.1-pro, \(\mathcal{S}_{\mathrm{direct}}\), episode & 22.3 [10.7, 34.0] & 12/0/3 \\
gpt-5.5 - gemini-3.1-pro, \(\mathcal{S}_{\mathrm{prog}}\), episode & 15.3 [6.3, 24.3] & 11/1/3 \\
gpt-5.5 - gemini-3.1-pro, \(\mathcal{S}_{\mathrm{prog}}\), held-out & 8.7 [1.9, 15.5] & 12/0/3 \\
gpt-5.5 - claude-opus-4.6, \(\mathcal{S}_{\mathrm{direct}}\), episode & 16.0 [6.7, 25.0] & 12/0/3 \\
gpt-5.5 - claude-opus-4.6, \(\mathcal{S}_{\mathrm{prog}}\), episode & 15.3 [6.9, 24.0] & 13/1/1 \\
gpt-5.5 - claude-opus-4.6, \(\mathcal{S}_{\mathrm{prog}}\), held-out & 5.3 [-2.5, 12.9] & 9/0/6 \\
gpt-5.5 - minimax-m2.7, \(\mathcal{S}_{\mathrm{direct}}\), episode & 58.7 [46.7, 70.7] & 15/0/0 \\
gpt-5.5 - minimax-m2.7, \(\mathcal{S}_{\mathrm{prog}}\), episode & 47.7 [36.7, 58.6] & 15/0/0 \\
gpt-5.5 - minimax-m2.7, \(\mathcal{S}_{\mathrm{prog}}\), held-out & 31.6 [21.6, 41.6] & 15/0/0 \\
\addlinespace[3pt]
\multicolumn{3}{l}{\textbf{Condition effects (closed-weight-agent averaged; minimax-m2.7 excluded)}}\\
Closed-weight-agent average: low - high noise, \(\mathcal{S}_{\mathrm{direct}}\), episode & 18.7 [8.8, 28.7] & 12/2/1 \\
Closed-weight-agent average: low - high noise, \(\mathcal{S}_{\mathrm{prog}}\), episode & 13.3 [4.7, 21.9] & 10/2/3 \\
Closed-weight-agent average: low - high noise, \(\mathcal{S}_{\mathrm{prog}}\), heldout & 13.5 [7.2, 19.9] & 13/0/2 \\
Closed-weight-agent average: \(\mathcal{S}_{\mathrm{direct}}-\mathcal{S}_{\mathrm{prog}}\), low noise, episode & 9.1 [4.7, 13.3] & 13/1/1 \\
Closed-weight-agent average: \(\mathcal{S}_{\mathrm{direct}}-\mathcal{S}_{\mathrm{prog}}\), high noise, episode & 3.8 [-2.7, 10.5] & 8/2/5 \\
\addlinespace[3pt]
\multicolumn{3}{l}{\textbf{Matched ablations}}\\
gpt-5.5: with - without text, \(\mathcal{S}_{\mathrm{direct}}\), episode & 32.0 [18.3, 45.7] & 13/2/0 \\
gpt-5.5: with - without text, \(\mathcal{S}_{\mathrm{prog}}\), episode & 28.0 [13.3, 42.5] & 12/1/2 \\
gpt-5.5: with - without text, \(\mathcal{S}_{\mathrm{prog}}\), held-out & 23.2 [14.1, 32.3] & 15/0/0 \\
gpt-5.5: 3-shot - 0-shot, \(\mathcal{S}_{\mathrm{direct}}\), episode & -5.3 [-16.7, 6.7] & 5/4/6 \\
gpt-5.5: 3-shot - 0-shot, \(\mathcal{S}_{\mathrm{prog}}\), episode & -11.3 [-26.0, 3.3] & 4/4/7 \\
gpt-5.5: 3-shot - 0-shot, \(\mathcal{S}_{\mathrm{prog}}\), held-out & 4.1 [-4.5, 11.9] & 12/0/3 \\
gemini-3.1-pro: with - without text, \(\mathcal{S}_{\mathrm{direct}}\), episode & 24.7 [8.6, 40.0] & 11/2/2 \\
gemini-3.1-pro: with - without text, \(\mathcal{S}_{\mathrm{prog}}\), episode & 20.0 [3.3, 36.0] & 11/2/2 \\
gemini-3.1-pro: with - without text, \(\mathcal{S}_{\mathrm{prog}}\), held-out & 11.3 [3.7, 18.9] & 10/0/5 \\
gemini-3.1-pro: 3-shot - 0-shot, \(\mathcal{S}_{\mathrm{direct}}\), episode & -8.7 [-28.3, 11.4] & 5/2/8 \\
gemini-3.1-pro: 3-shot - 0-shot, \(\mathcal{S}_{\mathrm{prog}}\), episode & -10.7 [-27.5, 5.0] & 5/3/7 \\
gemini-3.1-pro: 3-shot - 0-shot, \(\mathcal{S}_{\mathrm{prog}}\), held-out & -0.4 [-9.5, 8.7] & 7/0/8 \\
\bottomrule
\end{tabularx}
\endgroup

\end{table*}
\endgroup

\clearpage
\ifdefined\tracebenchDocsCombinedAppendixMode
\def\tracebenchAppendixPromptExampleCombinationsStart{%
  \docTitle{Appendix: Example Prompts Across Experimental Settings}%
  \phantomsection\label{docstart:appendix_prompt_example_combinations}%
  \section*{Example Prompts Across Experimental Settings}%
}
\def\tracebenchAppendixPromptExampleCombinationsEnd{}
\else\ifdefined\tracebenchCombinedAppendixMode
\def\tracebenchAppendixPromptExampleCombinationsStart{%
  \section{Example Prompts Across Experimental Settings}%
  \label{app:prompt-example-combinations}%
}
\def\tracebenchAppendixPromptExampleCombinationsEnd{}
\else
\documentclass[11pt]{article}
\usepackage{styles/manual_docs_style}

\def\tracebenchAppendixPromptExampleCombinationsStart{%
  \begin{document}%
  \docTitle{Appendix: Example Prompts Across Experimental Settings}%
  \phantomsection\label{docstart:appendix_prompt_example_combinations}%
  \section*{Example Prompts Across Experimental Settings}%
}
\def\tracebenchAppendixPromptExampleCombinationsEnd{\end{document}}
\fi\fi
\tracebenchAppendixPromptExampleCombinationsStart

For concreteness, all examples use \texttt{BallDrop}.
The subsections show the six prompt templates used across the main experiment and the two ablations, organized by domain context, submission mode, and availability of labeled examples.

\subsection*{With Domain Context, Direct Answer, No Labeled Examples}
\begin{DocCode}
The time series in the test_samples/ folder were generated by simulating a 1D vertical point-mass ball model under gravity.
While the ball is above the ground, its motion is governed by gravity and quadratic air drag acting opposite the direction of travel, and the position evolves according to the current velocity.
Ground interaction is modeled with a restitution-based hard-stop law.
When the ball reaches the ground with impact speed at or above 0.5 m/s, the impact is treated as instantaneous and the post-impact speed is set by the coefficient of restitution e in [0,1].
The rebound points upward and its magnitude equals e times the pre-impact speed.
When the impact speed is below the threshold, the ball does not rebound and instead enters static contact at the ground.

Observed Signals:
col1: ball height
col2: ball velocity
col3: contact impulse (N*s)
col4: time

For each simulation, either no parameter changes occur, or exactly one parameter among the allowed labels changes during the observed simulation interval.
If a parameter changes, it undergoes a single instantaneous step change at an unknown time during the observed interval.

Allowed labels:
["coefficient of restitution", "mass", "drag coefficient", "gravity acceleration", "no parameter change"]

Use "no parameter change" if there is no evidence in the data of a parameter change during the observed interval.

Task:
Create a file named results.json in the current working directory.
For each file in test_samples/, return a ranked list of labels.
The first label is your final top-1 prediction and should be the single label you think is most likely correct.
You may include additional labels only when the evidence is genuinely ambiguous.
Additional labels are treated as lower-confidence alternatives.
The output must be valid JSON with exactly this structure:
{

  "<filename_1>": ["<top_1_label>"],
  "<filename_2>": ["<top_1_label>", "<optional_lower_confidence_label>"]

}

Requirements:
- Include one entry for every Parquet file in test_samples/.
- Every returned label must exactly match one of the allowed labels.
- The order of labels matters: the first label is the top-1 prediction.
- Do not include duplicate labels for a sample.
- Do not return all labels unless the evidence is genuinely ambiguous across all labels.

Evaluation:
The primary evaluation metric is top-1 accuracy: the first label in each returned list is compared with the hidden correct label.
A secondary shortlist score may also be reported. For a returned list of length m, the sample receives score 1/m if the hidden correct label appears anywhere in the list, and 0 otherwise. Therefore, unnecessary extra labels reduce the secondary score.

Additional requirements:
- If you create intermediate files, images, scripts, or notes while solving the task, create them in the current working directory.
- Internet access is disabled.
\end{DocCode}

\subsection*{With Domain Context, Programmatic Submission, No Labeled Examples}
\begin{DocCode}
Context:
The time series in the test_samples/ folder were generated by simulating a 1D vertical point-mass ball model under gravity.
While the ball is above the ground, its motion is governed by gravity and quadratic air drag acting opposite the direction of travel, and the position evolves according to the current velocity.
Ground interaction is modeled with a restitution-based hard-stop law.
When the ball reaches the ground with impact speed at or above 0.5 m/s, the impact is treated as instantaneous and the post-impact speed is set by the coefficient of restitution e in [0,1].
The rebound points upward and its magnitude equals e times the pre-impact speed.
When the impact speed is below the threshold, the ball does not rebound and instead enters static contact at the ground.

Observed Signals:
col1: ball height
col2: ball velocity
col3: contact impulse (N*s)
col4: time

For each simulation, either no parameter changes occur, or exactly one parameter among the allowed labels changes during the observed simulation interval.
If a parameter changes, it undergoes a single instantaneous step change at an unknown time during the observed interval.

Allowed labels:
["coefficient of restitution", "mass", "drag coefficient", "gravity acceleration", "no parameter change"]

Use "no parameter change" if there is no evidence in the data of a parameter change during the observed interval.

Task:
Create a Python script named rule.py in the current working directory.
The script must define exactly this function:
def predict(df) -> list[str]:
The input df is a pandas DataFrame containing one sample with columns col1, col2, col3, and col4.
The function will be called on samples inside test_samples/ and on additional held-out samples with the same schema and label set.
For each dataframe, predict(df) must return a ranked list of labels. The first label is the final top-1 prediction and should be the single label most likely to be correct. Additional labels are optional lower-confidence alternatives and should only be included when the evidence is genuinely ambiguous.

Requirements for predict(df):
- Return a Python list of strings.
- Every returned label must exactly match one of the allowed labels.
- The first returned label is the top-1 prediction.
- Do not include duplicate labels.
- Do not return all labels unless the evidence is genuinely ambiguous across all labels.
- You may inspect any file while developing rule.py, but the final submitted rule.py must not read, open, import, or depend on any files at prediction time, including test_samples/
- The final rule.py must be able to run on a dataframe alone.

Evaluation:
The primary evaluation metric is top-1 accuracy: the first label returned by predict(df) is compared with the hidden correct label.
A secondary shortlist score may also be reported. For a returned list of length m, the sample receives score 1/m if the hidden correct label appears anywhere in the list, and 0 otherwise. Therefore, unnecessary extra labels reduce the secondary score.

Additional requirements:
- If you create intermediate files, images, scripts, or notes while solving the task, create them in the current working directory.
- Internet access is disabled.
\end{DocCode}

\subsection*{With Domain Context, Direct Answer, Three Labeled Examples per Class}
\begin{DocCode}
The time series in the test_samples/ and train_samples/ folders were generated by simulating a 1D vertical point-mass ball model under gravity.
While the ball is above the ground, its motion is governed by gravity and quadratic air drag acting opposite the direction of travel, and the position evolves according to the current velocity.
Ground interaction is modeled with a restitution-based hard-stop law.
When the ball reaches the ground with impact speed at or above 0.5 m/s, the impact is treated as instantaneous and the post-impact speed is set by the coefficient of restitution e in [0,1].
The rebound points upward and its magnitude equals e times the pre-impact speed.
When the impact speed is below the threshold, the ball does not rebound and instead enters static contact at the ground.

Observed Signals:
col1: ball height
col2: ball velocity
col3: contact impulse (N*s)
col4: time

For each simulation, either no parameter changes occur, or exactly one parameter among the allowed labels changes during the observed simulation interval.
If a parameter changes, it undergoes a single instantaneous step change at an unknown time during the observed interval.

Allowed labels:
["coefficient of restitution", "mass", "drag coefficient", "gravity acceleration", "no parameter change"]

Use "no parameter change" if there is no evidence in the data of a parameter change during the observed interval.

Task:
Create a file named results.json in the current working directory.
For each file in test_samples/, return a ranked list of labels.
The first label is your final top-1 prediction and should be the single label you think is most likely correct.
You may include additional labels only when the evidence is genuinely ambiguous.
Additional labels are treated as lower-confidence alternatives.
The output must be valid JSON with exactly this structure:
{

  "<filename_1>": ["<top_1_label>"],
  "<filename_2>": ["<top_1_label>", "<optional_lower_confidence_label>"]

}

To help with this task, you can use the labeled `train_samples/` directory. The corresponding labels are available in `train_labels.json` file.

Requirements:
- Include one entry for every Parquet file in test_samples/.
- Every returned label must exactly match one of the allowed labels.
- The order of labels matters: the first label is the top-1 prediction.
- Do not include duplicate labels for a sample.
- Do not return all labels unless the evidence is genuinely ambiguous across all labels.

Evaluation:
The primary evaluation metric is top-1 accuracy: the first label in each returned list is compared with the hidden correct label.
A secondary shortlist score may also be reported. For a returned list of length m, the sample receives score 1/m if the hidden correct label appears anywhere in the list, and 0 otherwise. Therefore, unnecessary extra labels reduce the secondary score.

Additional requirements:
- If you create intermediate files, images, scripts, or notes while solving the task, create them in the current working directory.
- Internet access is disabled.
\end{DocCode}

\subsection*{With Domain Context, Programmatic Submission, Three Labeled Examples per Class}
\begin{DocCode}
Context:
The time series in the test_samples/ and train_samples/ folders were generated by simulating a 1D vertical point-mass ball model under gravity.
While the ball is above the ground, its motion is governed by gravity and quadratic air drag acting opposite the direction of travel, and the position evolves according to the current velocity.
Ground interaction is modeled with a restitution-based hard-stop law.
When the ball reaches the ground with impact speed at or above 0.5 m/s, the impact is treated as instantaneous and the post-impact speed is set by the coefficient of restitution e in [0,1].
The rebound points upward and its magnitude equals e times the pre-impact speed.
When the impact speed is below the threshold, the ball does not rebound and instead enters static contact at the ground.

Observed Signals:
col1: ball height
col2: ball velocity
col3: contact impulse (N*s)
col4: time

For each simulation, either no parameter changes occur, or exactly one parameter among the allowed labels changes during the observed simulation interval.
If a parameter changes, it undergoes a single instantaneous step change at an unknown time during the observed interval.

Allowed labels:
["coefficient of restitution", "mass", "drag coefficient", "gravity acceleration", "no parameter change"]

Use "no parameter change" if there is no evidence in the data of a parameter change during the observed interval.

Task:
Create a Python script named rule.py in the current working directory.
The script must define exactly this function:
def predict(df) -> list[str]:
The input df is a pandas DataFrame containing one sample with columns col1, col2, col3, and col4.
The function will be called on samples inside test_samples/ and on additional held-out samples with the same schema and label set.
For each dataframe, predict(df) must return a ranked list of labels. The first label is the final top-1 prediction and should be the single label most likely to be correct. Additional labels are optional lower-confidence alternatives and should only be included when the evidence is genuinely ambiguous.

To help with this task, you can use the labeled train_samples/ directory while developing rule.py. The corresponding labels are available in train_labels.json.

Requirements for predict(df):
- Return a Python list of strings.
- Every returned label must exactly match one of the allowed labels.
- The first returned label is the top-1 prediction.
- Do not include duplicate labels.
- Do not return all labels unless the evidence is genuinely ambiguous across all labels.
- You may inspect train_samples/ and train_labels.json while developing rule.py, but the final submitted rule.py must not read, open, import, or depend on any files at prediction time, including train_samples/, test_samples/ or train_labels.json
- The final rule.py must be able to run on a dataframe alone.

Evaluation:
The primary evaluation metric is top-1 accuracy: the first label returned by predict(df) is compared with the hidden correct label.
A secondary shortlist score may also be reported. For a returned list of length m, the sample receives score 1/m if the hidden correct label appears anywhere in the list, and 0 otherwise. Therefore, unnecessary extra labels reduce the secondary score.

Additional requirements:
- If you create intermediate files, images, scripts, or notes while solving the task, create them in the current working directory.
- Internet access is disabled.
\end{DocCode}

\subsection*{No Domain Context, Direct Answer, Three Labeled Examples per Class}
\begin{DocCode}
Context:
The time series in the test_samples/ and train_samples/ folders were generated by a simulator of an unknown physical phenomenon.
The column meanings are unknown, except for the last column, which represents time.

For each simulation, either no parameter changes occur, or exactly one parameter corresponding to one of the allowed changes during the observed simulation interval.
If a parameter changes, it undergoes a single instantaneous step change at an unknown time during the observed interval.

["label_0", "label_1", "label_2", "label_3"] denote different parameter changes, while "label_4" denotes that no parameter changed.

Task:
Create a file named results.json in the current working directory.
For each file in test_samples/, return a ranked list of labels.
The first label is your final top-1 prediction and should be the single label you think is most likely correct.
You may include additional labels only when the evidence is genuinely ambiguous.
Additional labels are treated as lower-confidence alternatives.
The output must be valid JSON with exactly this structure:
{

  "<filename_1>": ["<top_1_label>"],
  "<filename_2>": ["<top_1_label>", "<optional_lower_confidence_label>"]

}

To help with this task, you can use the labeled `train_samples/` directory. The corresponding labels are available in `train_labels.json` file.

Requirements:
- Include one entry for every Parquet file in test_samples/.
- Every returned label must exactly match one of the allowed labels.
- The order of labels matters: the first label is the top-1 prediction.
- Do not include duplicate labels for a sample.
- Do not return all labels unless the evidence is genuinely ambiguous across all labels.

Evaluation:
The primary evaluation metric is top-1 accuracy: the first label in each returned list is compared with the hidden correct label.
A secondary shortlist score may also be reported. For a returned list of length m, the sample receives score 1/m if the hidden correct label appears anywhere in the list, and 0 otherwise. Therefore, unnecessary extra labels reduce the secondary score.

Additional requirements:
- If you create intermediate files, images, scripts, or notes while solving the task, create them in the current working directory.
- Internet access is disabled.
\end{DocCode}

\subsection*{No Domain Context, Programmatic Submission, Three Labeled Examples per Class}
\begin{DocCode}
Context:
The time series in the test_samples/ and train_samples/ folders were generated by a simulator of an unknown physical phenomenon.
The column meanings are unknown, except for the last column, which represents time.

For each simulation, either no parameter changes occur, or exactly one parameter corresponding to one of the allowed changes during the observed simulation interval.
If a parameter changes, it undergoes a single instantaneous step change at an unknown time during the observed interval.

["label_0", "label_1", "label_2", "label_3"]  denote different parameter changes, while "label_4" denotes that no parameter changed.

Task:
Create a Python script named rule.py in the current working directory.
The script must define exactly this function:
def predict(df) -> list[str]:
The input df is a pandas DataFrame containing one sample with columns col1, col2, col3, and col4.
The function will be called on samples inside test_samples/ and on additional held-out samples with the same schema and label set.
For each dataframe, predict(df) must return a ranked list of labels. The first label is the final top-1 prediction and should be the single label most likely to be correct. Additional labels are optional lower-confidence alternatives and should only be included when the evidence is genuinely ambiguous.

To help with this task, you can use the labeled train_samples/ directory while developing rule.py. The corresponding labels are available in train_labels.json.

Requirements for predict(df):
- Return a Python list of strings.
- Every returned label must exactly match one of the allowed labels.
- The first returned label is the top-1 prediction.
- Do not include duplicate labels.
- Do not return all labels unless the evidence is genuinely ambiguous across all labels.
- You may inspect train_samples/ and train_labels.json while developing rule.py, but the final submitted rule.py must not read, open, import, or depend on any files at prediction time, including train_samples/, test_samples/ or train_labels.json
- The final rule.py must be able to run on a dataframe alone.

Evaluation:
The primary evaluation metric is top-1 accuracy: the first label returned by predict(df) is compared with the hidden correct label.
A secondary shortlist score may also be reported. For a returned list of length m, the sample receives score 1/m if the hidden correct label appears anywhere in the list, and 0 otherwise. Therefore, unnecessary extra labels reduce the secondary score.

Additional requirements:
- If you create intermediate files, images, scripts, or notes while solving the task, create them in the current working directory.
- Internet access is disabled.
\end{DocCode}

\tracebenchAppendixPromptExampleCombinationsEnd

\clearpage
\setlength{\floatsep}{6pt plus 2pt minus 2pt}
\setlength{\textfloatsep}{8pt plus 2pt minus 2pt}
\setlength{\intextsep}{6pt plus 2pt minus 2pt}

\def\TraceBenchKDDTableLead{%
  \section{Per-simulator results tables}%
  \label{app:per-simulator-result-tables}%
  \par\medskip
}

\begin{table*}[!tbp]
\ifdefined\TraceBenchKDDTableLead
\TraceBenchKDDTableLead
\global\let\TraceBenchKDDTableLead\relax
\fi
\centering
\caption{Per-\termSimulator results for the main experiment under \(\kappa_{\mathrm{low}}\) and submission mode \(\mathcal{S}_{\mathrm{direct}}\).}
\label{tab:app-per-simulator-gentle-01234-cloud}
\begingroup
\normalsize
\providecommand{\scorepm}[2]{{\normalsize $#1 \pm #2$}}
\setlength{\tabcolsep}{2.5pt}
\setlength{\aboverulesep}{0pt}
\setlength{\belowrulesep}{2pt}
\renewcommand{\arraystretch}{1.02}
\begin{tabularx}{\textwidth}{@{}>{\raggedright\arraybackslash}X*{3}{>{\centering\arraybackslash}p{0.215\textwidth}}@{}}
\toprule
\multicolumn{4}{c}{\textbf{Episode Accuracy}} \\*
\midrule
 & \texttt{BallDrop} & \texttt{BounceBall} & \texttt{MassSlide} \\*
\midrule
\texttt{gpt-5.5} & {\bfseries\boldmath \scorepm{1.000}{0.000}} & \scorepm{0.840}{0.089} & {\bfseries\boldmath \scorepm{0.960}{0.055}} \\*
\texttt{gemini-3.1-pro} & \scorepm{0.680}{0.179} & {\bfseries\boldmath \scorepm{0.920}{0.084}} & \scorepm{0.680}{0.084} \\*
\texttt{claude-opus-4.6} & \scorepm{0.820}{0.130} & \scorepm{0.920}{0.130} & \scorepm{0.820}{0.110} \\*
\texttt{minimax-m2.7} & \scorepm{0.160}{0.114} & \scorepm{0.180}{0.164} & \scorepm{0.540}{0.152} \\
\bottomrule
\end{tabularx}
\endgroup

\end{table*}

\begin{table*}[!tbp]
\centering
\caption{Per-\termSimulator results for the main experiment under \(\kappa_{\mathrm{low}}\) and submission mode \(\mathcal{S}_{\mathrm{prog}}\).}
\label{tab:app-per-simulator-gentle-01234-flame}
\begingroup
\normalsize
\providecommand{\scorepm}[2]{{\normalsize $#1 \pm #2$}}
\setlength{\tabcolsep}{2.5pt}
\setlength{\aboverulesep}{0pt}
\setlength{\belowrulesep}{2pt}
\renewcommand{\arraystretch}{1.02}
\begin{tabularx}{\textwidth}{@{}>{\raggedright\arraybackslash}X*{3}{>{\centering\arraybackslash}p{0.215\textwidth}}@{}}
\toprule
\multicolumn{4}{c}{\textbf{Episode Accuracy}} \\*
\midrule
 & \texttt{BallDrop} & \texttt{BounceBall} & \texttt{MassSlide} \\*
\midrule
\texttt{gpt-5.5} & {\bfseries\boldmath \scorepm{0.820}{0.239}} & {\bfseries\boldmath \scorepm{0.860}{0.089}} & {\bfseries\boldmath \scorepm{0.880}{0.084}} \\*
\texttt{gemini-3.1-pro} & \scorepm{0.720}{0.217} & \scorepm{0.720}{0.164} & \scorepm{0.760}{0.089} \\*
\texttt{claude-opus-4.6} & \scorepm{0.620}{0.164} & \scorepm{0.680}{0.148} & \scorepm{0.760}{0.055} \\*
\texttt{minimax-m2.7} & \scorepm{0.180}{0.130} & \scorepm{0.180}{0.084} & \scorepm{0.480}{0.179} \\
\midrule
\addlinespace[5pt]
\multicolumn{4}{c}{\textbf{Held-out Accuracy}} \\*
\midrule
 & \texttt{BallDrop} & \texttt{BounceBall} & \texttt{MassSlide} \\*
\midrule
\texttt{gpt-5.5} & {\bfseries\boldmath \scorepm{0.722}{0.147}} & {\bfseries\boldmath \scorepm{0.728}{0.077}} & \scorepm{0.564}{0.070} \\*
\texttt{gemini-3.1-pro} & \scorepm{0.606}{0.123} & \scorepm{0.578}{0.163} & \scorepm{0.629}{0.124} \\*
\texttt{claude-opus-4.6} & \scorepm{0.612}{0.068} & \scorepm{0.534}{0.128} & {\bfseries\boldmath \scorepm{0.697}{0.059}} \\*
\texttt{minimax-m2.7} & \scorepm{0.223}{0.044} & \scorepm{0.165}{0.018} & \scorepm{0.474}{0.034} \\
\bottomrule
\end{tabularx}
\endgroup

\end{table*}

\begin{table*}[!tbp]
\centering
\caption{Per-\termSimulator results for the main experiment under \(\kappa_{\mathrm{high}}\) and submission mode \(\mathcal{S}_{\mathrm{direct}}\).}
\label{tab:app-per-simulator-gentle-01234-pine}
\begingroup
\normalsize
\providecommand{\scorepm}[2]{{\normalsize $#1 \pm #2$}}
\setlength{\tabcolsep}{2.5pt}
\setlength{\aboverulesep}{0pt}
\setlength{\belowrulesep}{2pt}
\renewcommand{\arraystretch}{1.02}
\begin{tabularx}{\textwidth}{@{}>{\raggedright\arraybackslash}X*{3}{>{\centering\arraybackslash}p{0.215\textwidth}}@{}}
\toprule
\multicolumn{4}{c}{\textbf{Episode Accuracy}} \\*
\midrule
 & \texttt{BallDrop} & \texttt{BounceBall} & \texttt{MassSlide} \\*
\midrule
\texttt{gpt-5.5} & {\bfseries\boldmath \scorepm{0.840}{0.261}} & {\bfseries\boldmath \scorepm{0.780}{0.179}} & {\bfseries\boldmath \scorepm{0.880}{0.179}} \\*
\texttt{gemini-3.1-pro} & \scorepm{0.440}{0.230} & \scorepm{0.500}{0.324} & \scorepm{0.740}{0.114} \\*
\texttt{claude-opus-4.6} & \scorepm{0.500}{0.187} & \scorepm{0.440}{0.219} & \scorepm{0.840}{0.134} \\*
\texttt{minimax-m2.7} & \scorepm{0.200}{0.158} & \scorepm{0.200}{0.122} & \scorepm{0.500}{0.235} \\
\bottomrule
\end{tabularx}
\endgroup

\end{table*}

\begin{table*}[!tbp]
\centering
\caption{Per-\termSimulator results for the main experiment under \(\kappa_{\mathrm{high}}\) and submission mode \(\mathcal{S}_{\mathrm{prog}}\).}
\label{tab:app-per-simulator-gentle-01234-orbit}
\begingroup
\normalsize
\providecommand{\scorepm}[2]{{\normalsize $#1 \pm #2$}}
\setlength{\tabcolsep}{2.5pt}
\setlength{\aboverulesep}{0pt}
\setlength{\belowrulesep}{2pt}
\renewcommand{\arraystretch}{1.02}
\begin{tabularx}{\textwidth}{@{}>{\raggedright\arraybackslash}X*{3}{>{\centering\arraybackslash}p{0.215\textwidth}}@{}}
\toprule
\multicolumn{4}{c}{\textbf{Episode Accuracy}} \\*
\midrule
 & \texttt{BallDrop} & \texttt{BounceBall} & \texttt{MassSlide} \\*
\midrule
\texttt{gpt-5.5} & {\bfseries\boldmath \scorepm{0.760}{0.207}} & {\bfseries\boldmath \scorepm{0.640}{0.321}} & \scorepm{0.800}{0.187} \\*
\texttt{gemini-3.1-pro} & \scorepm{0.460}{0.152} & \scorepm{0.500}{0.255} & \scorepm{0.680}{0.110} \\*
\texttt{claude-opus-4.6} & \scorepm{0.520}{0.130} & \scorepm{0.440}{0.305} & {\bfseries\boldmath \scorepm{0.820}{0.130}} \\*
\texttt{minimax-m2.7} & \scorepm{0.340}{0.167} & \scorepm{0.200}{0.071} & \scorepm{0.520}{0.249} \\
\midrule
\addlinespace[5pt]
\multicolumn{4}{c}{\textbf{Held-out Accuracy}} \\*
\midrule
 & \texttt{BallDrop} & \texttt{BounceBall} & \texttt{MassSlide} \\*
\midrule
\texttt{gpt-5.5} & {\bfseries\boldmath \scorepm{0.580}{0.165}} & {\bfseries\boldmath \scorepm{0.478}{0.169}} & \scorepm{0.584}{0.104} \\*
\texttt{gemini-3.1-pro} & \scorepm{0.399}{0.060} & \scorepm{0.352}{0.142} & \scorepm{0.569}{0.043} \\*
\texttt{claude-opus-4.6} & \scorepm{0.464}{0.105} & \scorepm{0.395}{0.190} & {\bfseries\boldmath \scorepm{0.639}{0.104}} \\*
\texttt{minimax-m2.7} & \scorepm{0.215}{0.030} & \scorepm{0.196}{0.045} & \scorepm{0.486}{0.064} \\
\bottomrule
\end{tabularx}
\endgroup

\end{table*}

\def\TraceBenchKDDTableLead{\subsection{Ablation Results}}

\begin{table*}[!tbp]
\ifdefined\TraceBenchKDDTableLead
\TraceBenchKDDTableLead
\global\let\TraceBenchKDDTableLead\relax
\fi
\centering
\caption{Per-\termSimulator results for the ablation without labeled examples under \(\kappa_{\mathrm{high}}\) and submission mode \(\mathcal{S}_{\mathrm{direct}}\).}
\label{tab:app-per-simulator-frost-01234-pine}
\begingroup
\normalsize
\providecommand{\scorepm}[2]{{\normalsize $#1 \pm #2$}}
\setlength{\tabcolsep}{2.5pt}
\setlength{\aboverulesep}{0pt}
\setlength{\belowrulesep}{2pt}
\renewcommand{\arraystretch}{1.02}
\begin{tabularx}{\textwidth}{@{}>{\raggedright\arraybackslash}X*{3}{>{\centering\arraybackslash}p{0.215\textwidth}}@{}}
\toprule
\multicolumn{4}{c}{\textbf{Episode Accuracy}} \\*
\midrule
 & \texttt{BallDrop} & \texttt{BounceBall} & \texttt{MassSlide} \\*
\midrule
\texttt{gpt-5.5} & {\bfseries\boldmath \scorepm{0.900}{0.122}} & {\bfseries\boldmath \scorepm{0.820}{0.148}} & {\bfseries\boldmath \scorepm{0.940}{0.134}} \\*
\texttt{gemini-3.1-pro} & \scorepm{0.760}{0.134} & \scorepm{0.440}{0.167} & \scorepm{0.740}{0.207} \\*
\bottomrule
\end{tabularx}
\endgroup

\end{table*}

\begin{table*}[!tbp]
\centering
\caption{Per-\termSimulator results for the ablation without labeled examples under \(\kappa_{\mathrm{high}}\) and submission mode \(\mathcal{S}_{\mathrm{prog}}\).}
\label{tab:app-per-simulator-frost-01234-orbit}
\begingroup
\normalsize
\providecommand{\scorepm}[2]{{\normalsize $#1 \pm #2$}}
\setlength{\tabcolsep}{2.5pt}
\setlength{\aboverulesep}{0pt}
\setlength{\belowrulesep}{2pt}
\renewcommand{\arraystretch}{1.02}
\begin{tabularx}{\textwidth}{@{}>{\raggedright\arraybackslash}X*{3}{>{\centering\arraybackslash}p{0.215\textwidth}}@{}}
\toprule
\multicolumn{4}{c}{\textbf{Episode Accuracy}} \\*
\midrule
 & \texttt{BallDrop} & \texttt{BounceBall} & \texttt{MassSlide} \\*
\midrule
\texttt{gpt-5.5} & {\bfseries\boldmath \scorepm{0.880}{0.045}} & {\bfseries\boldmath \scorepm{0.720}{0.164}} & {\bfseries\boldmath \scorepm{0.940}{0.134}} \\*
\texttt{gemini-3.1-pro} & \scorepm{0.640}{0.114} & \scorepm{0.680}{0.228} & \scorepm{0.640}{0.207} \\*
\midrule
\addlinespace[5pt]
\multicolumn{4}{c}{\textbf{Held-out Accuracy}} \\*
\midrule
 & \texttt{BallDrop} & \texttt{BounceBall} & \texttt{MassSlide} \\*
\midrule
\texttt{gpt-5.5} & {\bfseries\boldmath \scorepm{0.629}{0.042}} & \scorepm{0.411}{0.083} & {\bfseries\boldmath \scorepm{0.478}{0.092}} \\*
\texttt{gemini-3.1-pro} & \scorepm{0.474}{0.107} & {\bfseries\boldmath \scorepm{0.433}{0.167}} & \scorepm{0.427}{0.071} \\*
\bottomrule
\end{tabularx}
\endgroup

\end{table*}

\begin{table*}[!tbp]
\centering
\caption{Per-\termSimulator results for the ablation without domain context under \(\kappa_{\mathrm{high}}\) and submission mode \(\mathcal{S}_{\mathrm{direct}}\).}
\label{tab:app-per-simulator-gentle-01234-harbor}
\begingroup
\normalsize
\providecommand{\scorepm}[2]{{\normalsize $#1 \pm #2$}}
\setlength{\tabcolsep}{2.5pt}
\setlength{\aboverulesep}{0pt}
\setlength{\belowrulesep}{2pt}
\renewcommand{\arraystretch}{1.02}
\begin{tabularx}{\textwidth}{@{}>{\raggedright\arraybackslash}X*{3}{>{\centering\arraybackslash}p{0.215\textwidth}}@{}}
\toprule
\multicolumn{4}{c}{\textbf{Episode Accuracy}} \\*
\midrule
 & \texttt{BallDrop} & \texttt{BounceBall} & \texttt{MassSlide} \\*
\midrule
\texttt{gpt-5.5} & {\bfseries\boldmath \scorepm{0.600}{0.235}} & {\bfseries\boldmath \scorepm{0.400}{0.187}} & {\bfseries\boldmath \scorepm{0.540}{0.230}} \\*
\texttt{gemini-3.1-pro} & \scorepm{0.220}{0.045} & \scorepm{0.200}{0.071} & \scorepm{0.520}{0.239} \\*
\texttt{claude-opus-4.6} & - & - & - \\*
\texttt{minimax-m2.7} & - & - & - \\
\bottomrule
\end{tabularx}
\endgroup

\end{table*}

\begin{table*}[!tbp]
\centering
\caption{Per-\termSimulator results for the ablation without domain context under \(\kappa_{\mathrm{high}}\) and submission mode \(\mathcal{S}_{\mathrm{prog}}\).}
\label{tab:app-per-simulator-gentle-01234-tide}
\begingroup
\normalsize
\providecommand{\scorepm}[2]{{\normalsize $#1 \pm #2$}}
\setlength{\tabcolsep}{2.5pt}
\setlength{\aboverulesep}{0pt}
\setlength{\belowrulesep}{2pt}
\renewcommand{\arraystretch}{1.02}
\begin{tabularx}{\textwidth}{@{}>{\raggedright\arraybackslash}X*{3}{>{\centering\arraybackslash}p{0.215\textwidth}}@{}}
\toprule
\multicolumn{4}{c}{\textbf{Episode Accuracy}} \\*
\midrule
 & \texttt{BallDrop} & \texttt{BounceBall} & \texttt{MassSlide} \\*
\midrule
\texttt{gpt-5.5} & {\bfseries\boldmath \scorepm{0.340}{0.230}} & {\bfseries\boldmath \scorepm{0.440}{0.241}} & {\bfseries\boldmath \scorepm{0.580}{0.205}} \\*
\texttt{gemini-3.1-pro} & \scorepm{0.240}{0.114} & \scorepm{0.300}{0.212} & \scorepm{0.500}{0.158} \\*
\texttt{claude-opus-4.6} & - & - & - \\*
\texttt{minimax-m2.7} & - & - & - \\
\midrule
\addlinespace[5pt]
\multicolumn{4}{c}{\textbf{Held-out Accuracy}} \\*
\midrule
 & \texttt{BallDrop} & \texttt{BounceBall} & \texttt{MassSlide} \\*
\midrule
\texttt{gpt-5.5} & {\bfseries\boldmath \scorepm{0.249}{0.089}} & {\bfseries\boldmath \scorepm{0.250}{0.128}} & \scorepm{0.447}{0.073} \\*
\texttt{gemini-3.1-pro} & \scorepm{0.208}{0.009} & \scorepm{0.199}{0.056} & {\bfseries\boldmath \scorepm{0.575}{0.058}} \\*
\texttt{claude-opus-4.6} & - & - & - \\*
\texttt{minimax-m2.7} & - & - & - \\
\bottomrule
\end{tabularx}
\endgroup

\end{table*}

\FloatBarrier

\clearpage
\section{Representative gpt-5.5 BallDrop Low-Noise Direct Trajectory}
\label{app:gpt-5-5-ball-drop-direct-example}
We summarize a representative agent trajectory from the main low-noise, direct-answer condition.
The complete trajectory is provided in the ancillary file \path{tracebench_representative_agent_trajectory.pdf}.

\end{document}